\documentclass[10pt]{article} % For LaTeX2e
\usepackage[accepted]{tmlr}
\usepackage{amsmath,amsfonts,bm}

\def\eqref#1{equation~\ref{#1}}
\def\1{\bm{1}}

\def\vs{{\bm{s}}}

\DeclareMathAlphabet{\mathsfit}{\encodingdefault}{\sfdefault}{m}{sl}
\SetMathAlphabet{\mathsfit}{bold}{\encodingdefault}{\sfdefault}{bx}{n}

\usepackage{hyperref}
\usepackage{url}
\usepackage{graphicx}

\usepackage{booktabs}
\usepackage{tabularx}
\usepackage{array}

\usepackage{subcaption}
\usepackage{booktabs}
\usepackage{cleveref}
\usepackage{algorithm}    % 提供「算法浮动体」（类似 figure/table）
\usepackage{algpseudocode} % 伪代码语法（支持行号、if/else 等结构）
\usepackage{amsmath}      % 处理数学符号（如空集 ∅、下标）
\usepackage{caption}      % 可选：统一调整标题样式
\usepackage{xcolor}

\title{WISE: A Long-Horizon Agent in Minecraft with Why-Which Reasoning}

\author{
  \name Renmin Cheng
  \email chengrm@lamda.nju.edu.cn \\
  \addr The Hong Kong University of Science and Technology (Guangzhou)
  \AND
  \name Changhao Chen\thanks{Changhao Chen is with Intelligent Transportation Thrust and Artificial Intelligence Thrust, The Hong Kong University of Science and Technology (Guangzhou), Guangzhou 511453, China, and also with  the Division of Emerging Interdisciplinary Areas, The Hong Kong University of Science and Technology, Hong Kong SAR 999077, China} \thanks{Corresponding author}
  \email changhaochen@hkust-gz.edu.cn \\
  \addr The Hong Kong University of Science and Technology (Guangzhou) \\
  The Hong Kong University of Science and Technology
}

\def\month{09}  % Insert correct month for camera-ready version
\def\year{2026} % Insert correct year for camera-ready version
\def\openreview{\url{https://openreview.net/forum?id=T8HuiP3yM9}} % Insert correct link to OpenReview for camera-ready version

\begin{document}

\maketitle

\begin{abstract}
\label{sed:abs}

Rapid advances have been made in developing general-purpose embodied agents in environments like Minecraft through the adoption of LLM-augmented hierarchical approaches. Despite their promise, low-level controllers often become performance bottlenecks due to repeated execution failures. We argue that a key limitation is not only the lack of episodic memory, but also the decoupling of \textit{what-where-when} memory from \textit{which-why} reasoning.
To address this, we propose \textbf{WISE} (Which-Why Informed Semantic Explorer), a long-horizon agent framework with an enhanced low-level controller equipped with a  Semantic Affordance Event Graph (SAEG) that augments episodic memory with  semantic affordance structure linking observations to task relevance. Unlike prior work such as MrSteve, which relies on feature similarity for retrieval, WISE enables robust recall under viewpoint changes and supports opportunistic task reordering through  affordance-grounded task reasoning. Building on this memory, we propose an Opportunistic Task Scheduler that dynamically re-prioritizes subtasks when  task-relevant opportunities are detected. We further equip WISE with a multi-scale progressive exploration strategy to provide spatially comprehensive observations for downstream reasoning.
Experiments show that WISE substantially improves task success and efficiency on long-horizon sparse tasks, particularly in settings requiring adaptive decision-making.
\end{abstract}

\section{Introduction}
\label{sec:intro}

\begin{figure}[htb]
  \centering

  \includegraphics[width=1\linewidth]{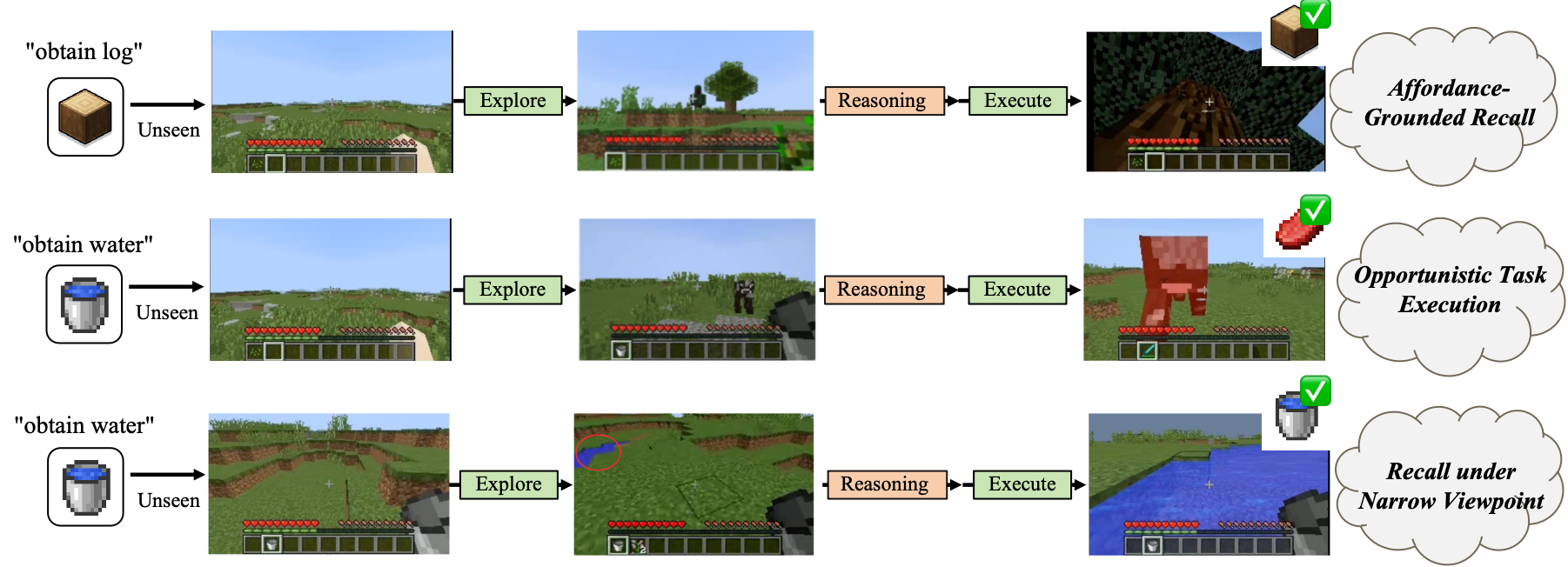} 
  \caption{Comparison of WISE and prior approaches across three key capabilities. \textbf{Affordance-Grounded Recall:} WISE recalls previously observed trees for the task ``obtain log''  by leveraging  the semantic affordance relation (Oak $\rightarrow$ log), whereas MrSteve relies solely on visual similarity. \textbf{Opportunistic Task Execution:} WISE dynamically reorders the task queue to exploit chance encounters (e.g., immediately killing a nearby cow), while prior approaches rigidly follow a predefined task sequence. \textbf{Robust Recall under Viewpoint Changes:} WISE employs semantic entity representations that are   less sensitive to viewpoint variations, enabling consistent recognition of the pond from different perspectives; prior approaches depend on raw visual features and often fail under viewpoint changes. Green check marks indicate successful behavior.}
  \label{fig:overview} 
\end{figure} 

Building generally capable embodied agents that can solve long-horizon tasks in complex, open-ended environments remains a fundamental challenge in artificial intelligence. Minecraft has emerged as a prominent benchmark for studying this problem because it presents a procedurally generated world with rich interaction dynamics requiring exploration, resource acquisition, tool crafting, and long-term planning \cite{guss2019minerl,fan2022minedojo,baker2022video}. Even seemingly simple objectives, such as \textit{``obtain beef''}, require multi-step reasoning: the agent must first infer the semantic link between beef and cows, then explore sparse biomes to locate cattle, navigate complex terrain, engage in interaction or combat, and finally collect the resulting resources. Failures at any stage (e.g., losing track of the target entity) often lead to costly re-exploration. More complex tasks, such as diamond acquisition, can require tens of thousands of environment steps \cite{lifshitz2023steve}, making reinforcement learning from scratch prohibitively difficult due to sparse rewards and vast exploration spaces. 

Recent progress has shown that Large Language Model (LLM)-augmented hierarchical frameworks provide a promising solution \cite{wang2023describe,zhou2024minedreamer}. These systems decompose long-horizon tasks into two levels: a high-level planner that leverages LLM reasoning to generate subgoals and a low-level controller that executes them. This decomposition largely reduces task complexity and has enabled notable advances in embodied decision making. For such frameworks to succeed, however, high-level planning and low-level execution must improve jointly. Existing research has primarily concentrated on enhancing high-level planning through skill libraries \cite{zhu2023ghost,wang2023describe,qin2024mp5}, multimodal experience repositories \cite{li2024optimus}, or increasingly powerful language reasoning modules. 

In contrast, comparatively little attention has been paid to the low-level controller, which often becomes the dominant bottleneck in practice \cite{cai2023open}. Existing methods frequently assume that once a subgoal is generated, the controller can reliably and efficiently execute it---an assumption that often breaks down in large, sparse environments. A recent attempt to address this limitation is MrSteve \cite{park2025mrsteve}, which augments Steve-1 with Place Event Memory (PEM). PEM organizes past observations according to \textit{what} happened, \textit{where} it occurred, and \textit{when} it was observed, enabling retrieval of previously encountered locations and events. For example, an agent may remember that it previously observed a cow in a forest region and later navigate back to that location.

However, we argue that the primary limitation is not only memory capacity itself, but also the absence of  task-relevant semantic structure over memory. Existing systems \cite{wang2024voyager, zhu2023ghost, park2025mrsteve} isolate episodic recall from decision making. PEM retrieves observations using cosine similarity between MineCLIP visual features and task embeddings, resulting in two key deficiencies. First, retrieval based on raw visual similarity is semantically brittle: changes in viewpoint, occlusion \cite{cai2023open, zhou2024minedreamer}, or environmental conditions frequently cause memory recall to fail. Second, PEM stores observations without representing   which task outcomes the observed entities afford. The memory may record that a cow was previously observed, but it cannot  represent why this observation matters---namely, that cows can provide beef. Consequently, when the future task becomes \textit{``obtain beef''}, the agent lacks a mechanism to identify previously encountered cows as relevant opportunities.

More fundamentally, existing controllers \cite{lifshitz2023steve, park2025mrsteve} execute subgoals according to fixed action sequences and cannot answer the question: \textit{which action should be performed right now given newly available information?} Consider an agent traveling to collect wood that unexpectedly encounters a cow. An ideal agent should immediately recognize that the encounter creates an opportunity to satisfy a future subgoal. Existing systems \cite{qin2024mp5, yuan2023skill, li2025larm} instead continue executing the prescribed sequence and return later, incurring unnecessary navigation cost. Exploration, memory, and planning operate as disconnected modules; consequently, the overall system  cannot coordinate newly acquired observations with memory retrieval and task ordering.

In this work, we introduce \textbf{WISE} (\textbf{W}hich-\textbf{W}hy \textbf{I}nformed \textbf{S}emantic \textbf{E}xplorer), a long-horizon agent framework that closes the loop between exploration, memory, and decision making. As illustrated in Figure~\ref{fig:overview}, WISE extends conventional episodic memory with  semantic affordance structure, enabling the agent not only to remember \textit{what}, \textit{where}, and \textit{when}, but also to understand \textit{why} an observation matters and \textit{which} future tasks it may enable.
The core of WISE is a  \textit{Semantic Affordance Event Graph}, a semantic memory structure that augments episodic observations with  semantic affordance relations extracted by a Vision-Language Model. Instead of storing observations as isolated visual memories, the graph links entities and downstream task outcomes through  affordance edges (e.g., \textit{cow} $\rightarrow$ \verb|CAN_OBTAIN| $\rightarrow$ \textit{beef}). This semantic layer enables  affordance-grounded retrieval and transforms memory from passive storage into actionable knowledge.
Building upon this representation, we introduce an \textit{Opportunistic Task Scheduler} that continuously reasons over  semantic affordance memories and dynamically reprioritizes pending subtasks. Rather than rigidly following a predefined execution sequence, the scheduler adapts to newly observed opportunities. When a  task-relevant entity appears, WISE immediately updates its task priorities and exploits the opportunity online, thereby avoiding redundant navigation and improving long-horizon efficiency.
Finally, to support this reasoning process with sufficiently rich observations, we introduce a multi-scale progressive exploration strategy that efficiently expands environmental coverage while minimizing revisitation. Together, exploration, memory, and scheduling form a closed-loop architecture: exploration acquires observations, memory converts observations into  semantic affordance knowledge, and planning consumes this knowledge for adaptive decision making.
Extensive experiments in large-scale Minecraft environments demonstrate the effectiveness of WISE. Compared with the current state-of-the-art low-level controller MrSteve \cite{park2025mrsteve}, WISE achieves a 14\% improvement in exploration coverage, a 30\% increase in sequential sparse task success with 26.4\% lower completion time, and a 44\% increase in adaptive non-sequential task success with 42.5\% less completion time. 
Ablation studies show that each component contributes to the overall effectiveness of WISE: removing the \textit{Semantic Affordance Event Graph}, \textit{Opportunistic Task Scheduler}, or \textit{Multi-scale Progressive Exploration} leads to a clear drop in performance.

Our contributions are as follows.
\begin{enumerate}
\item We identify a key but underexplored limitation in long-horizon embodied agents: the disconnect between episodic memory and  affordance-grounded decision-making. We argue that an effective low-level controller must go beyond remembering \textit{what}, \textit{where}, and \textit{when}, and instead reason about \textit{why} observations matter and \textit{which} future actions they enable.
\item We propose \textbf{WISE}, a unified embodied framework that closes the loop between exploration, memory, and planning. Its core is a    Semantic Affordance Event Graph that augments episodic memory with VLM-derived   semantic affordance relations, enabling robust retrieval and  affordance-grounded reasoning.
\item We introduce an Opportunistic Task Scheduler that dynamically reprioritizes subtasks based on   semantic affordance relevance, transforming the low-level controller from a rigid executor into an adaptive decision-maker capable of exploiting newly emerging opportunities.
\item We present a multi-scale progressive exploration strategy that improves coverage efficiency while providing the reasoning module with spatially comprehensive observations for long-horizon decision making.
\end{enumerate}

\section{Related work}
\label{sec:related_work}
\subsection{Low-Level Control in Minecraft}
Early works trained policy models for simple Minecraft tasks \cite{guss2019minerl}. VPT \cite{baker2022video} showed that vision-to-action mappings can be learned from unlabeled online videos at scale. Steve-1 \cite{lifshitz2023steve} extended VPT by conditioning on text instructions. GROOT \cite{caigroot} used reference videos instead of text for goal-conditioned control. MineDreamer \cite{zhou2024minedreamer} leveraged Steve-1 to generate subgoal images for better execution.

More recent research focuses on increasing the architectural complexity and adaptability of controllers. The STEVE Series \cite{zhao2024steve} systematically explored the interplay between planning granularity and observation encoding. Odyssey \cite{liu2025odyssey} and LARM \cite{li2025larm} introduced open-world skill libraries and auto-regressive models to bridge the gap between reactive control and long-horizon deliberative planning. ADAM \cite{yuadam2025} and Steve-Evolving \cite{xie2026steve} represent the cutting edge of this direction: the former constructs a causal graph through interaction, while the latter distills execution diagnoses into reusable guardrails and skills. However, these systems often treat memory as a high-level knowledge store rather than an integrated component of low-level execution. MrSteve \cite{park2025mrsteve} introduced Place Event Memory (PEM) to provide the low-level controller with episodic context, but its retrieval relies on raw visual similarity. WISE addresses this by extending PEM into a  Semantic Affordance Event Graph, enabling the controller to reason about why a memory is relevant to future tasks.

\subsection{Memory in Agents}
Memory is essential for long-horizon tasks, and many studies have explored different storage and retrieval mechanisms for embodied agents. \cite{sumers2023cognitive} propose a framework that categorizes agent memory into three types: in-context, external, and in-weights. They highlight the trade-off between retrieval accuracy and update speed, which directly motivates the two-level retrieval design in WISE.

At the plan level, several memory systems have been developed. Voyager \cite{wang2024voyager} stores reusable skill programs across episodes, enabling lifelong learning in Minecraft without human intervention. GITM \cite{zhu2023ghost} uses structured world knowledge to improve goal decomposition for technology-tree tasks. JARVIS-1 \cite{wang2025jarvis} and MP5 \cite{qin2024mp5} store both multimodal observations and successful plans for situation-aware retrieval, with JARVIS-1 showing that multimodal goal specification greatly improves retrieval precision. Optimus-1 \cite{li2024optimus} introduces a hybrid memory pool that summarizes visual, textual, and action trajectories, achieving strong results on long-horizon benchmarks.

Despite these advances, most existing memory systems focus on high-level planning---they store which plan worked, not why a specific low-level observation matters. MrSteve's PEM is the first memory system designed for low-level control, but it still retrieves memories based on raw visual features. WISE fills this gap by constructing a  Semantic Affordance Event Graph that connects low-level observations to high-level goals through explicit  semantic affordance edges. A related system, \cite{yuadam2025} also builds a causal graph via interaction, but it targets high-level knowledge accumulation and does not integrate causal memory with a dynamic low-level scheduler.

\subsection{Exploration in Open Worlds}

Exploration strategies span several paradigms, each with a characteristic failure mode.  Count-based methods \cite{tang2017exploration,chang2024goat} encourage novelty but accumulate coverage gaps super-linearly with map size. Frontier-based methods \cite{yamauchi1998frontier, 10999073} improve boundary awareness but ignore large unvisited interior regions. NovelCraft \cite{feeney2023novelcraft} further showed that existing exploration policies fail to detect and adapt to novel objects in procedurally generated worlds, underscoring the need for multi-scale hierarchical planning that balances global coverage and local detail.
VPT-Nav \cite{park2025mrsteve} adapts to local dynamics but is reactive rather than globally planned.  Active information-gain methods \cite{chaplotlearning,chaplot2020neural,zhang2021noveld} are theoretically well-motivated but computationally intractable under Minecraft's $\leq50$ ms per-step real-time constraint. WISE's multi-scale strategy is the first to synthesise all three strengths in Minecraft:  $O(\log n)$ region lookup and target selection from quadtree indexing, boundary-awareness from frontier scoring, and a local completeness guarantee from Voronoi decomposition.

\section{Method}
\label{sec:method}

\begin{figure}[htb]
  \centering
  \setlength{\belowcaptionskip}{-10pt}
  \includegraphics[width=0.85\linewidth]{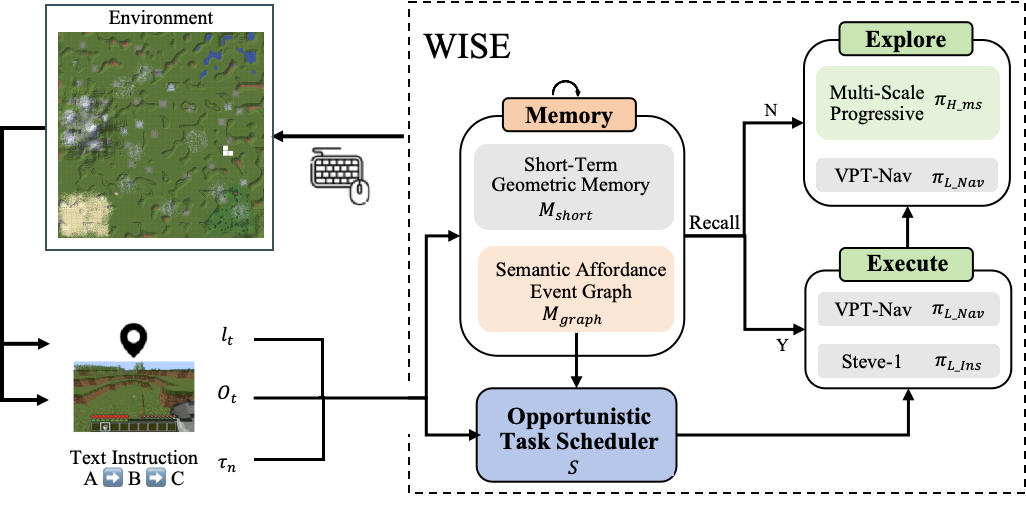} 
  \caption{Overall architecture of WISE. Given a text instruction or subtask sequence (e.g., ``A $\rightarrow$ B $\rightarrow$ C''), the agent alternates among three phases: \textit{Explore}, \textit{Reason}, and \textit{Execute}. During exploration, a multi-scale progressive policy systematically explores the environment and stores observations in two memory systems: a short-term geometric memory for efficient retrieval of recent observations and a  Semantic Affordance Event Graph for storing semantic entities, spatial instances, and semantic affordance relations. During execution, the agent uses VPT-Nav for goal-directed navigation and Steve-1 for low-level instruction execution. The Opportunistic Task Scheduler dynamically reorders subtasks according to  semantic affordance memories retrieved from SAEG.}
  \label{fig:wise_architecture} 
\end{figure}

In this section, we introduce WISE, a memory-augmented embodied agent framework for sparse long-horizon tasks in Minecraft. We first describe the problem setting and then present the construction of each module.

\textbf{ Problem setting.}
We consider sparse long-horizon task scenarios in which an embodied agent continuously receives a sequence of tasks $\{\tau_n\}_{n=1}^{\infty}$ via textual instructions (e.g., \textit{``Obtain beef''}) generated either directly by the environment or decomposed by a large language model (LLM). We assume that task-relevant resources occur infrequently and are sparsely distributed throughout large environments. Consequently, successful task completion requires the agent not only to memorize observations from explored regions, but also to semantically understand how these observations may be relevant to future tasks.

At every timestep $t$, the agent receives an observation
\begin{equation}
X_t = \{O_t,l_t,t\},
\end{equation}
where $O_t=i_t \in  \mathbb{R}^{H_{\text{img}}\times W_{\text{img}}\times C_{\text{img}}}$ denotes the first-person RGB observation.  Here, $H_{\text{img}}$, $W_{\text{img}}$, and $C_{\text{img}}$ denote the image height, image width, and number of color channels, respectively. The pose vector
\begin{equation}
l_t=(coord_x,coord_y,coord_z,yaw,pitch)\in\mathbb{R}^{5}
\end{equation}
contains the agent's 3D coordinates and camera orientation relative to the initial frame of reference.

\textbf{Framework overview.}
WISE is a memory-augmented agent framework consisting of three major components: a \textit{Memory Module}, a \textit{Solver Module}, and an \textit{Opportunistic Task Scheduler}. The memory module $M_t$ contains two complementary memory structures:
\begin{equation}
M_t=(M_{\mathrm{short}},M_{\mathrm{SAEG}}).
\end{equation}
First,  $M_{\mathrm{SAEG}}$ is a \textit{Semantic Affordance Event Graph} (SAEG) that extends conventional episodic memory with semantic entities, spatial instances, and task-relevant semantic affordance relations. Second, a \textit{short-term geometric memory}, $M_{\mathrm{short}}$, provides low-latency retrieval of recently observed experiences.

Based on retrieved memory contents, a mode selector within the solver module dynamically switches between \textit{Explore} and \textit{Execute} modes. Together with a multi-scale progressive exploration policy, these modules form the complete WISE framework. \Cref{alg:wise_loop} outlines the task-solving loop of WISE.  By jointly exploring, memorizing, and using semantic affordance relations for retrieval and scheduling, WISE enables efficient decision-making for sparse long-horizon tasks. SAEG guides target selection and task scheduling; the retrieved locations are passed to VPT-Nav as navigation targets, and Steve-1 executes the selected natural-language subgoal.

 For the decision loop, we denote the pending subtask queue at timestep $t$ as $Q_t=[\tau_1,\ldots,\tau_m]$. For each pending subtask $\tau_i$, $\mathcal{C}_i$ denotes the retrieved memory candidate set. A stored memory candidate $x$ consists of a visual embedding $e_x$, a stored pose $l_x$, and a timestamp $t_x$.

\begin{algorithm}[htbp]
\caption{WISE Single Decision Loop}
\label{alg:wise_loop}
\begin{algorithmic}[1]
\Require Current observation $X_t=(O_t,l_t,t)$; memory $M_t=(M_{\mathrm{short}},M_{\mathrm{SAEG}})$; pending subtask queue $Q_t=[\tau_1,\ldots,\tau_m]$; scheduler $S$.
\State Update $M_{\mathrm{short}}$ with the current observation $X_t$.
\State Enqueue eligible keyframes for asynchronous SAEG update.
\State Incorporate available VLM outputs into $M_{\mathrm{SAEG}}$ after grounding and filtering.
\For{each pending subtask $\tau_i \in Q_t$}
    \State Retrieve memory candidates $\mathcal{C}_i \leftarrow \textsc{Retrieve}(M_{\mathrm{short}},M_{\mathrm{SAEG}},\tau_i)$ using Eq.~\ref{eq:retrieval-score}.
    \State Compute priority $p_i \leftarrow S(\tau_i,\mathcal{C}_i,l_t)$ using Eq.~\ref{eq:priority}.
\EndFor
\State Select $\tau^\star=\arg\max_{\tau_i\in Q_t}p_i$.
\State Select $x^\star=\textsc{TopCandidate}(\mathcal{C}_{\tau^\star})$.
\If{$x^\star \neq \emptyset$}
    \State Set navigation target $l^\star \leftarrow l_{x^\star}$.
    \State Navigate to $l^\star$ with $\pi_{\mathrm{L\text{-}Nav}}$.
    \State Execute subtask $\tau^\star$ with $\pi_{\mathrm{Inst}}$.
    \State Remove $\tau^\star$ from $Q_t$ if completed.
\Else
    \State Select an exploration target using the multi-scale explorer.
    \State Explore with $\pi_{\mathrm{H\text{-}MS}}$ and $\pi_{\mathrm{L\text{-}Nav}}$.
\EndIf
\end{algorithmic}
\end{algorithm}

 Algorithm~\ref{alg:wise_loop} makes explicit the three operations performed by the online controller: memory update, task-level prioritization, and low-level execution. VLM inference is not part of the blocking control path. The online loop only selects candidate keyframes and appends them to a pending VLM buffer; graph construction is performed asynchronously as described in Section~\ref{subsec:vlm_saeg}. At each timestep, WISE queries the latest available version of $M_{\mathrm{SAEG}}$. If no valid retrieved memory exists for the selected subtask, the agent switches to exploration.

\subsection{ Semantic Affordance Event Graph for Episodic Memory}
\label{subsec:saeg}

Episodic memory in embodied agents typically records \textit{what} happened, \textit{where} it happened, and \textit{when} it occurred. However, for long-horizon tasks, this information alone is insufficient. An agent must additionally understand \textit{why} a past observation may matter for a current or future objective, and \textit{which} pending subtask should be executed when a useful opportunity appears.

The proposed  Semantic Affordance Event Graph extends this paradigm by explicitly encoding  semantic affordance relationships  among entities and tasks. This transforms episodic experiences into structured semantic knowledge that supports opportunistic reasoning and dynamic replanning.
%  We introduce the Semantic Affordance Event Graph (SAEG) to address this limitation. SAEG augments episodic memory by linking observed entities, their spatial locations, and the task-relevant outcomes that these entities can enable under Minecraft mechanics. We define a \textit{semantic affordance relation} as a task-relevant link between an observed entity and an obtainable resource or executable outcome. For example, a \textit{cow} is linked to \textit{beef}, and a \textit{tree} is linked to \textit{logs}. These relations provide the \textit{why} dimension of memory: an observation is useful not only because of where and when it was seen, but also because of which future task it may support.

\subsubsection{Short-Term Geometric Memory}

\begin{figure}[htb]
  \centering
  \setlength{\belowcaptionskip}{-10pt}
  \includegraphics[width=1\linewidth]{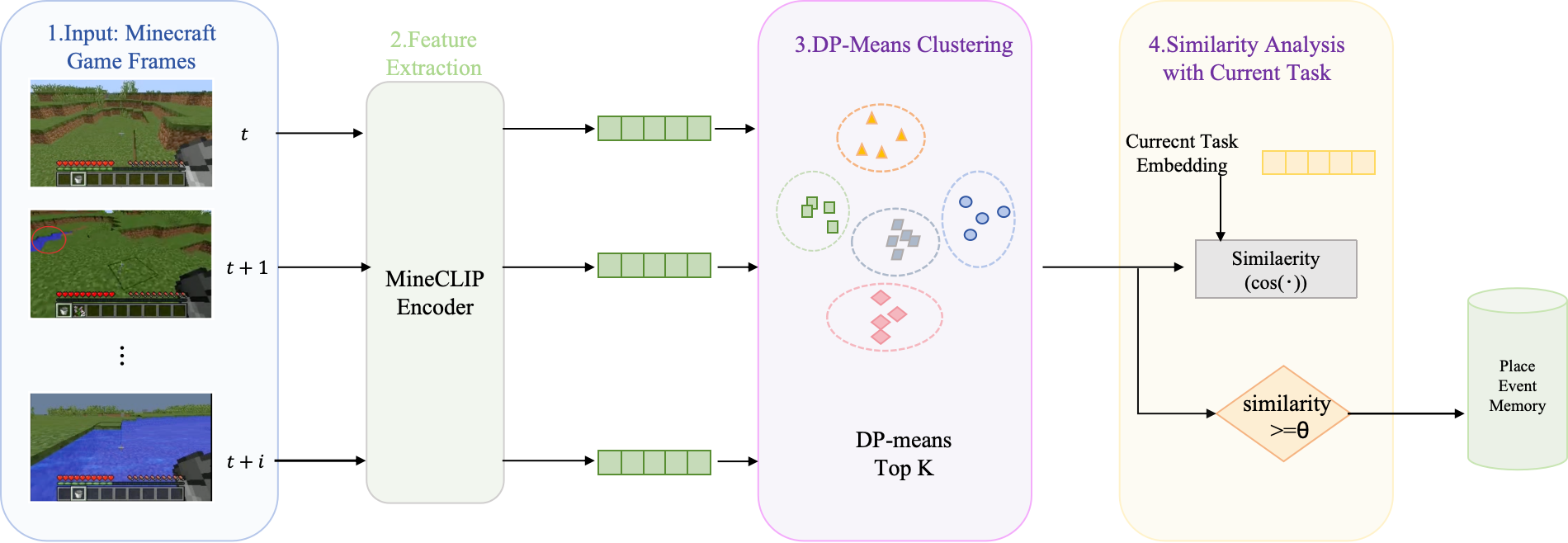} 
  \caption{Short-Term Geometric Memory. Given Minecraft video frames, a MineCLIP encoder extracts visual embeddings, which are grouped into event clusters using DP-Means clustering. The system then computes cosine similarity between the current task embedding and each cluster centroid. Clusters with similarity above a predefined threshold are ranked, and the top-$K$ matches are retrieved as memory cues.}
  \label{fig:short_memory} 
\end{figure}

As shown in Figure~\ref{fig:short_memory}, given Minecraft game frames, a MineCLIP encoder first extracts visual representations. DP-Means~\cite{dinari2022revisiting} clustering is then applied to group temporally related observations into event clusters. Subsequently, cosine similarity between the current task embedding and each event-cluster center is computed. Clusters whose similarity exceeds a predefined threshold are ranked, and the top-$K$ clusters are retrieved as memory cues. This mechanism follows the memory design adopted in MrSteve, but  does not explicitly represent entity-level semantic affordances or task-enabling relations.

To enable rapid and low-latency retrieval of recent experiences, WISE maintains a short-term memory structure identical to PEM. Observation embeddings are computed as
\begin{equation}
e_t =  Enc_{\text{vis}}(i_{t-T_{\text{win}}:t}),
\end{equation}
where  $Enc_{\text{vis}}$ denotes the MineCLIP visual encoder and $T_{\text{win}}=16$ denotes the temporal window size. Experience frames are represented as
\begin{equation}
x_t=\{e_t,l_t,t\},
\end{equation}
where $e_t$ is the stored visual embedding, $l_t$ is the pose, and $t$ is the timestep.

The stored frames are organized into place clusters based on the agent's 2D position and yaw orientation. Within each place cluster, event clusters are further formed via DP-Means clustering on video embeddings. When memory capacity is exceeded, the oldest frame from the largest event cluster is removed. This memory layer efficiently answers \textit{where} and \textit{when} queries. However, its representation of \textit{what} remains restricted to visual similarity: it captures appearance but not task-relevant semantic affordances.

\subsubsection{\textit{ VLM-Driven Semantic Affordance Graph Construction}}
\label{subsec:vlm_saeg}

\begin{figure}[htb!]
  \centering
  \setlength{\belowcaptionskip}{-10pt}
  \includegraphics[width=1\linewidth]{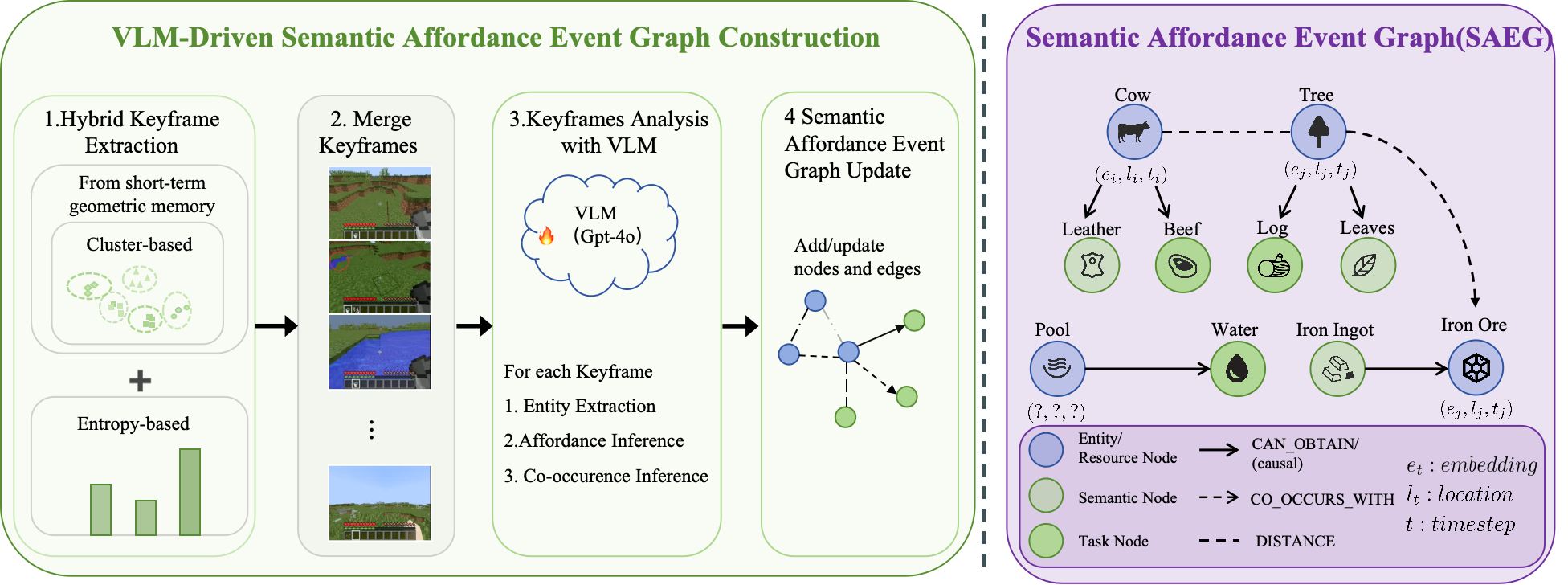} 
  \caption{VLM-Driven construction of the Semantic Affordance Event Graph. Hybrid keyframe extraction first selects representative observations from short-term geometric memory using complementary clustering-based and entropy-based strategies, followed by redundancy reduction through keyframe merging. A Vision-Language Model then analyzes each keyframe to extract semantic entities,  identify semantic affordance relations (e.g., ``cow $\rightarrow$ beef''), and identify spatial co-occurrence patterns. The resulting information is used to update  SAEG by inserting or modifying entity nodes, spatial instances, semantic affordance edges, and co-occurrence edges.}
  \label{fig:graph} 
\end{figure}

To understand why a particular observation matters, we construct a semantic affordance event graph on top of the short-term geometric memory. As shown in Figure~\ref{fig:graph}, graph construction begins with keyframe selection. Informative keyframes should preserve both semantic content and visual diversity from the original observation stream. We therefore use a hybrid keyframe extraction strategy that combines clustering-based and entropy-based selection.

Specifically, one set of keyframes is obtained from feature clusters in $M_{\mathrm{short}}$, preserving representative observations across different scenes and events. In parallel, an image entropy criterion selects visually informative frames with potentially rich semantic content. The two sets are merged and filtered to remove near-duplicate frames, producing a compact and semantically diverse set of candidate keyframes.

To prevent VLM inference from interrupting real-time control, WISE decouples graph construction from action execution. The selected keyframes are first placed into a pending VLM buffer instead of being processed immediately in the control loop. An asynchronous VLM worker periodically batches the buffered keyframes into groups of 16 and sends one streaming request for each group. This design amortizes the cost of VLM inference over multiple keyframes while allowing the agent to continue interacting with the environment.

During this process, the agent continues to act using the short-term geometric memory and the latest available version of SAEG. Once a VLM response is returned, the parsed outputs are grounded and filtered, after which SAEG is updated in a synchronized step by inserting or updating entity nodes, spatial attributes, timestamps, semantic affordance relations, and co-occurrence relations. Therefore, the online controller never waits for VLM inference, while the scheduler always queries the most recent updated graph.

The number of selected keyframes is adaptive and depends on the explored trajectory and redundancy filtering. The experimental section reports the average number of VLM requests per episode. The exact VLM prompt template, structured output schema, grounding vocabulary, and filtering rules are provided in Appendix~A.

For each keyframe, the VLM performs three operations.

\textbf{Entity extraction.}
The VLM identifies semantically meaningful elements in the scene, including animals, resources, terrain structures, and environmental objects. These entities are added as graph nodes and associated with the spatial coordinates and timestamp of the originating observation.

\textbf{Semantic affordance relation extraction.} For each extracted entity, the VLM identifies candidate entity--outcome relations, which are then grounded and validated against the Minecraft task-resource vocabulary. For example, a \textit{cow} entity is connected to \textit{beef} and \textit{leather} via \texttt{CAN\_OBTAIN} relations, while a \textit{tree} is connected to \textit{logs}. These edges encode the \textit{why} dimension of memory: observing a cow is useful for the task ``obtain beef'' because the cow provides an actionable semantic affordance for that task.

\textbf{Spatial co-occurrence relation extraction.}
For entities observed in the same keyframe or nearby spatial locations, WISE adds or updates \texttt{CO\_OCCURS\_WITH} edges. These edges encode spatial regularities among entities and help retrieve nearby resources or landmarks during exploration and task execution.

After VLM processing, extracted entity names are normalized and grounded to the Minecraft vocabulary and task-resource database. Duplicate entity instances are merged when they have the same normalized entity type and nearby spatial coordinates. Semantic affordance edges are retained only when the source entity is observed in the keyframe and the target resource or outcome is valid under the task-resource vocabulary. Together, these operations transform low-level episodic observations into structured semantic affordance memory suitable for retrieval and scheduling.

\subsubsection{\textit{Two-Level Retrieval with Semantic Affordance Augmentation}}
\label{subsec:two_level_retrieval}

When the agent receives a new task $\tau_n$, memory retrieval proceeds at two complementary levels. First, the short-term geometric layer computes cosine similarity between the task embedding $z_{\tau_n}=Enc_{\text{text}}(\tau_n)$ and stored observation embeddings, returning the top-$k$ visually matched memories.

Second, the semantic layer queries the Semantic Affordance Event Graph. The task description is matched against entity or resource nodes, after which semantic affordance edges are traversed to identify observations that are task-relevant. For example, a query such as \textit{``obtain beef''} first matches the \textit{beef} node and then follows incoming \texttt{CAN\_OBTAIN} edges to associated \textit{cow} instances, thereby retrieving their stored spatial locations. This semantic affordance retrieval process explains why a memory is selected: a cow observation is retrieved for a beef-related task not because its visual appearance resembles beef, but because the graph explicitly encodes their semantic affordance  relationship.

The outputs from both retrieval pathways are integrated using a weighted scoring function:
\begin{equation}
Score(x,\tau_n)=
\lambda_{\mathrm{ret}}\cdot sim(e_x, z_{\tau_n})
+
(1-\lambda_{\mathrm{ret}})\cdot AffordanceMatch(x,\tau_n)
\label{eq:retrieval-score}
\end{equation}

where $x$ is a stored memory candidate, $e_x$ is the visual embedding of candidate $x$, $z_{\tau_n}=Enc_{\text{text}}(\tau_n)$ is the task embedding, and $sim(\cdot,\cdot)$ denotes cosine similarity. $AffordanceMatch(x,\tau_n)$ equals 1 if candidate $x$ contains an entity connected to task $\tau_n$ through a semantic affordance relation in $M_{\mathrm{SAEG}}$, and 0 otherwise. $\lambda_{\mathrm{ret}}\in[0,1]$ controls the trade-off between visual similarity and semantic affordance matching.

Candidates returned by either the visual pathway or the SAEG pathway are merged and re-ranked by Eq.~\ref{eq:retrieval-score}. The highest-ranked candidate, if available, is selected as the navigation target.

This formulation provides a semantic affordance retrieval pathway in addition to visual similarity retrieval. When visual appearance changes due to viewpoint variation, occlusion, or lighting, SAEG can still retrieve a relevant memory through entity--task affordance links. The retrieved location is then passed to the navigation module as a target; SAEG guides target selection and task scheduling rather than directly modifying the low-level policy input.

\subsection{Opportunistic Task Scheduler for Adaptive Task Selection}
\label{subsec:scheduler}

Given multiple pending subtasks, continuously evolving observations, and semantic affordance memories, the central question becomes: \textit{which} subtask should be executed at the current moment?

Existing systems such as MrSteve rely on a fixed task sequence generated by a high-level planner. The low-level controller executes subgoals strictly according to the prescribed order and cannot deviate even when the environment presents opportunities that would make future subtasks easier to accomplish. This rigid strategy often results in inefficient behavior and missed opportunities. For example, suppose an agent encounters a cow while traveling to collect wood. Under a fixed execution order, the agent must first finish the wood-collection task and later return to the same location to obtain beef, introducing unnecessary navigation cost.

WISE addresses this limitation with an Opportunistic Task Scheduler that dynamically prioritizes subtasks according to the current observation and retrieved semantic affordance memories. At each decision step, the scheduler constructs a unified decision context from three information sources: the current visual observation $O_t$, the pending subgoal queue $Q_t=[\tau_1,\ldots,\tau_m]$, and the retrieved memory candidate sets $\{\mathcal{C}_i\}_{i=1}^{m}$. Each pending subgoal $\tau_i$ is associated with a memory cue consisting of its highest-ranked retrieved observation, corresponding spatial coordinates, and the semantic affordance relations connecting the memory to the task.

Each pending subgoal $\tau_i$ receives a composite priority score:
\begin{equation}
\text{Priority}(\tau_i,t)=
w_u U(\tau_i,t)
+
w_a A(\tau_i,t)
+
w_n \left[1-N(\tau_i,t)\right]
\label{eq:priority}
\end{equation}

where $U(\tau_i,t)$ is the urgency score, $A(\tau_i,t)$ is the semantic affordance relevance score, and $N(\tau_i,t)$ is the normalized navigation cost. All three terms are normalized to $[0,1]$.

The urgency term preserves the preference of the original task queue. In our implementation, it is computed from the rank of $\tau_i$ in the pending queue $Q_t$:
\begin{equation}
U(\tau_i,t)=1-\frac{r_i-1}{\max(|Q_t|-1,1)},
\label{eq:urgency}
\end{equation}
where $r_i$ is the rank of $\tau_i$ in $Q_t$. Thus, earlier subtasks receive higher default urgency.

The semantic affordance relevance term is computed from the latest available SAEG state and the retrieved candidate set $\mathcal{C}_i$:
\begin{equation}
A(\tau_i,t)=
\begin{cases}
\max_{x\in \mathcal{C}_i} AffordanceMatch(x,\tau_i), & \mathcal{C}_i \neq \emptyset,\\
0, & \mathcal{C}_i = \emptyset.
\end{cases}
\label{eq:affordance_rel}
\end{equation}
This term becomes high when a retrieved memory contains an entity connected to $\tau_i$ through a semantic affordance relation. Since SAEG updates are asynchronous, newly observed opportunities become available to the scheduler once their corresponding VLM outputs have been incorporated into SAEG. Before that, the agent continues to rely on short-term geometric memory and exploration.

The navigation cost term is defined as
\begin{equation}
N(\tau_i,t)=
\min\left(
\frac{d(l_t,l_i^\star)}{d_{\max}},
1
\right),
\label{eq:navcost}
\end{equation}
where $l_i^\star$ is the location of the top retrieved candidate for $\tau_i$, $d(l_t,l_i^\star)$ denotes the distance estimate used by the navigation module, and $d_{\max}$ is a map-scale normalization constant. If no valid candidate is retrieved for $\tau_i$, we set $N(\tau_i,t)=1$.

For instance, when the graph contains the semantic affordance relation
\[
\textit{cow} \rightarrow \texttt{CAN\_OBTAIN} \rightarrow \textit{beef},
\]
the semantic affordance relevance score for the task ``obtain beef'' becomes high; otherwise, it remains near zero.

We set $w_u=0.3$, $w_a=0.5$, and $w_n=0.2$ in the main experiments, reflecting the design choice that semantic affordance relevance should strongly influence opportunistic task switching. After scoring, the scheduler reorders the pending task queue according to descending priority. If the highest-priority task differs from the currently active one, execution is preempted and control switches to the new task.

This strategy is opportunistic because it exploits favorable environmental encounters that fixed plans ignore. Task switching is driven by semantic affordance reasoning: when SAEG links an observed or remembered entity to a pending subgoal, the scheduler can raise the priority of that subgoal and execute it earlier.

\textbf{Illustrative example.}
Consider an agent executing the task sequence:
\[
[\texttt{find water} \rightarrow \texttt{collect logs} \rightarrow \texttt{obtain beef}].
\]

Once a cow observation has been incorporated into SAEG, the corresponding semantic affordance relation becomes available to the scheduler:
\[
\textit{cow} \rightarrow \texttt{CAN\_OBTAIN} \rightarrow \textit{beef}.
\]

The scheduler then recomputes task priorities, causing the semantic affordance relevance of ``obtain beef'' to increase.
The reordered execution queue becomes:
\[
[\texttt{obtain beef} \rightarrow \texttt{find water} \rightarrow \texttt{collect logs}].
\]
The agent immediately acquires beef and subsequently resumes the remaining objectives, avoiding a future return trip. By contrast, a fixed-sequence policy such as MrSteve would ignore the encountered cow and revisit the same location later.

\subsection{Multi-Scale Progressive Exploration}
\label{subsec:exploration}

The effectiveness of SAEG and the Opportunistic Task Scheduler depends on access to sufficiently diverse and spatially complete observations.
When no task-relevant memory can be retrieved and no exploitable opportunity is identified by the scheduler, the agent must actively explore the environment to acquire new information.

To address this challenge, WISE employs a \textit{multi-scale progressive exploration} strategy designed to cover large environments while minimizing redundant exploration. The strategy consists of three hierarchical tiers operating at different spatial resolutions: global region selection, regional frontier refinement, and local completion. Each tier addresses limitations left unresolved by the preceding level.

\textbf{Global exploration tier.}
At the coarsest level, the global exploration tier determines which large-scale region should be visited next. We partition the environment using a quadtree representation. Each quadtree node $N_i$ represents a candidate macro-region. 
Let $P_t$ denote the agent's current 2D map position. We define the global exploration score as
\begin{equation}
Score_G(N_i)=\alpha_G \cdot Depth(N_i)-\beta_G \cdot Dist(P_t,N_i),
\label{eq:global-score}
\end{equation}
where $Depth(N_i)$ measures the normalized depth or unexplored granularity of region $N_i$, $Dist(P_t,N_i)$ is the normalized distance from the current position to the region, and $\alpha_G,\beta_G$ are global-tier weights. The first term encourages the agent to visit fine-grained unexplored regions, while the second term penalizes distant targets.

The quadtree tier provides efficient $O(\log n)$ region lookup and next-target selection, where $n$ is the number of candidate spatial regions. This complexity describes the cost of selecting the next macro-region from the spatial index. The realized coverage still depends on the low-level controller, terrain geometry, and the timestep budget.

\textbf{Regional exploration tier.}
After arriving at a selected macro-region, the agent switches to regional exploration. Frontier points are defined as boundaries between explored and unexplored cells. 
Each frontier point $f_i$ is scored as
\begin{equation}
Score_R(f_i)=
\alpha_R \cdot Cover(f_i)
+
\beta_R \cdot Novel(f_i)
-
\gamma_R \cdot Dist(P_t,f_i),
\label{eq:regional-score}
\end{equation}
where $Cover(f_i)$ estimates the uncovered area visible behind frontier $f_i$, $Novel(f_i)$ measures the information novelty of the frontier, $Dist(P_t,f_i)$ is the normalized distance from the agent to the frontier, and $\alpha_R,\beta_R,\gamma_R$ are regional-tier weights. The frontier with the highest score is selected as the next local exploration target. This tier expands exploration boundaries and compensates for the fact that global quadtree selection alone does not specify how to cover a selected region.

\textbf{Local completion tier.}
Although global and regional exploration expand the explored area, small interior gaps may remain. 
To target these residual gaps, WISE applies a local completion stage based on Voronoi decomposition. Let $U_i$ denote a residual unexplored region identified from the local map. We score each such region by
\begin{equation}
Score_L(U_i)=
\alpha_L \cdot Area(U_i)
-
\beta_L \cdot Dist(P_t,U_i),
\label{eq:local-score}
\end{equation}
where $Area(U_i)$ is the normalized area of the residual unexplored region, $Dist(P_t,U_i)$ is the normalized distance from the current position to that region, and $\alpha_L,\beta_L$ are local-tier weights. The agent moves toward the highest-scoring residual region to improve local coverage completion.

Together, these three levels form a coarse-to-fine exploration framework. The global tier improves map-level target selection, the regional tier expands exploration boundaries, and the local tier improves coverage completion by targeting residual gaps. Through this progressive design, WISE acquires diverse observations required for semantic memory construction and semantic affordance reasoning in sparse long-horizon environments.

\section{Experiments}
\label{sec:exper}
In this section, we comprehensively evaluate WISE across four complementary dimensions. Section~4.1 describes the experimental protocol and evaluation setup. Section~4.2 evaluates large-scale exploration performance to assess the effectiveness of the proposed multi-scale exploration strategy. Section~4.3 evaluates sparse sequential (ABA-Sparse) and Section~4.4 evaluates non-sequential (ABC-Sparse) task completion, jointly evaluating memory retrieval and adaptive planning capabilities. Section~4.5 presents controlled ablation studies to isolate the contribution of each component. Additional implementation details are provided in Appendix A. All experiments are conducted in Minecraft Java Edition through the MineRL platform.

\subsection{Experiment Setups}

\textbf{Hardware and software environment.} All experiments are conducted on a single server equipped with four NVIDIA RTX 4080 16GB GPUs running Ubuntu 22.04 LTS, with each rollout using one GPU. The software stack includes Minecraft Java Edition v1.11.2, MineRL v0.4.4, PyTorch v2.5.1, and the official MineCLIP checkpoint from MineDojo. GPT-4o (\texttt{gpt-4o-2024-05-13}) is used as the Vision-Language Model for semantic affordance graph construction.

\textbf{Baseline Selection.}  We compare WISE against the following representative baselines:
\begin{itemize}
    \item \textit{Steve-1}~\cite{lifshitz2023steve} is the most widely adopted low-level instruction-following controller built upon the Video Pre-Training (VPT) framework~\cite{baker2022video}. Following prior work~\cite{caigroot,zhou2024minedreamer}, we additionally provide Steve-1 with the instruction ``Go Explore'' to assess its exploration capability.
    \item \textit{MrSteve}~\cite{park2025mrsteve} is the current state-of-the-art low-level controller based on Place Event Memory (PEM). It stores \textit{what--where--when} information and retrieves memories using visual similarity. MrSteve additionally employs a hierarchical exploration strategy consisting of a count-based high-level goal selector and a VPT-Nav low-level controller.
    \item \textit{ WISE (SAEG only)} is an ablation variant that retains the proposed  SAEG while removing multi-scale exploration and the Opportunistic Task Scheduler. This variant isolates the contribution of semantic affordance memory without adaptive planning or enhanced exploration.
    \item  \textit{VLM Entity Retrieval without SAEG} uses GPT-4o-generated entity labels for retrieval but does not construct semantic affordance edges or perform graph traversal.
    \item \textit{WISE (Quadtree-Based)} is another ablation variant that uses only the global quadtree exploration tier while removing frontier-based regional exploration and Voronoi-based local completion.
\end{itemize}

We explicitly exclude API-privileged agents, including Voyager~\cite{wang2024voyager}, JARVIS-1~\cite{wang2025jarvis}, and GITM~\cite{zhu2023ghost}, because these methods directly access environment states such as entity coordinates, inventory information, and terrain metadata. In contrast, WISE and all evaluated baselines rely exclusively on raw visual observations. Direct comparison across these settings would confound architectural capability with information privilege and therefore lead to misleading conclusions.

\textbf{Evaluation Metrics.}
We adopt three complementary evaluation metrics.
\begin{itemize}
    \item \textit{Map Coverage} measures the fraction of map cells visited or observed within the agent's field of view. Areas that remain unexplored cannot contribute observations for downstream memory or planning.
    \item \textit{Success Rate} denotes the percentage of episodes that successfully complete all subgoals within a fixed timestep budget.
    \item 
\textit{ Average Completion Timesteps} is computed only over successful episodes and measures task efficiency.
\item For  sparse sequential (ABA-Sparse)  tasks, we additionally report \textit{return localization steps}, defined as the number of steps required to relocate the original location of object A after collecting object B. This metric isolates memory retrieval performance from navigation ability.

\end{itemize}

\textbf{Core hyperparameters.} Key parameters are summarized below, while complete implementation details and selection procedures are provided in Appendix A.
For exploration, we set the quadtree depth weight to $\alpha_G=0.6$ and distance penalty to $\beta_G=0.4$. Frontier utility weights (coverage, novelty, and distance) and Voronoi utility weights (area and distance) are normalized.
For memory, the entropy threshold is $\theta_{\mathrm{ent}}=0.15$, cosine similarity threshold is $\theta_c=0.85$, and top-$k$ retrieval uses $k=10$.
For planning, urgency, semantic affordance relevance, and navigation cost weights are set to $w_u=0.3$, $w_a=0.5$, and $w_n=0.2$, respectively.
The maximum quadtree depth is set to 6 for $384\times384$ maps and 5 for $128\times128$ maps. Voronoi decomposition is updated every 40 steps, with the minimum unexplored area threshold set to 60 pixels.

\textbf{VLM cost and latency.} On average, WISE performs approximately 14 GPT-4o requests per episode. Each request consumes roughly 4,575 input tokens and 1,618 output tokens, resulting in an estimated API cost of \$0.48 per episode. Since VLM inference runs asynchronously, it does not block the online control loop.

\subsection{Large-Scale Exploration Performance}
We first evaluate the efficiency with which each method explores and covers the environment. The primary metric is map coverage, defined as the fraction of map cells visited or observedwithin the agent's field of view. Table~\ref{tab:coverage} reports coverage under varying step budgets.

\textbf{Simulated environment.} We first evaluate under an idealized movement model in which the agent moves deterministically by a fixed distance at each timestep (one block per step). This setting removes low-level execution variability and isolates the intrinsic spatial efficiency of the exploration strategy itself.
On the small $128\times128$ map with a budget of 6,000 steps, WISE achieves 99\% coverage, compared to 59\% for MrSteve and 93\% for the quadtree-based variant. On the larger $384\times384$ map with 20,000 steps, WISE reaches 98\% coverage whereas MrSteve plateaus at 83\%.
The advantage extends beyond final coverage. WISE explores at a rate of 0.097 map coverage per 1,000 steps, compared to 0.067 for MrSteve. Because movement stochasticity has been removed, these results reflect the intrinsic efficiency of the proposed multi-scale exploration strategy.

\begin{figure}[t]
    \centering
    \includegraphics[width=0.42\linewidth]{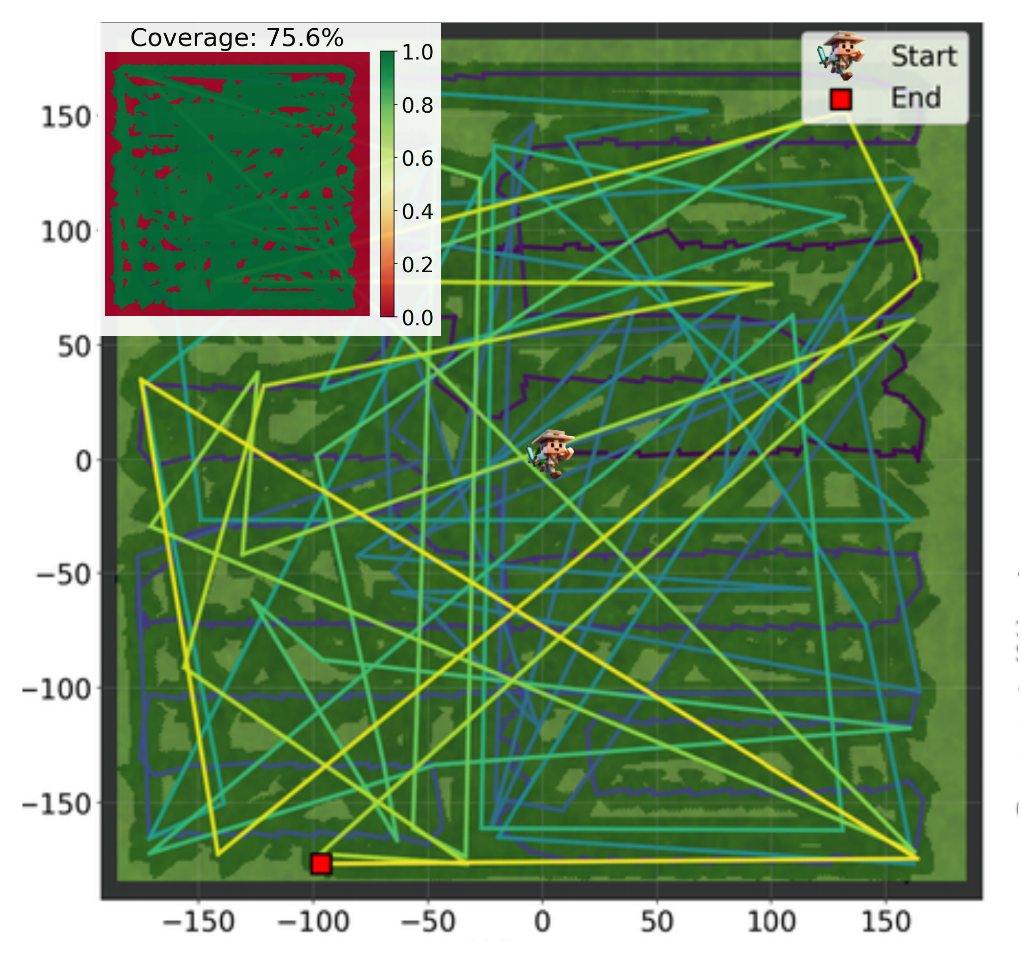} 
    \includegraphics[width=0.4\linewidth]{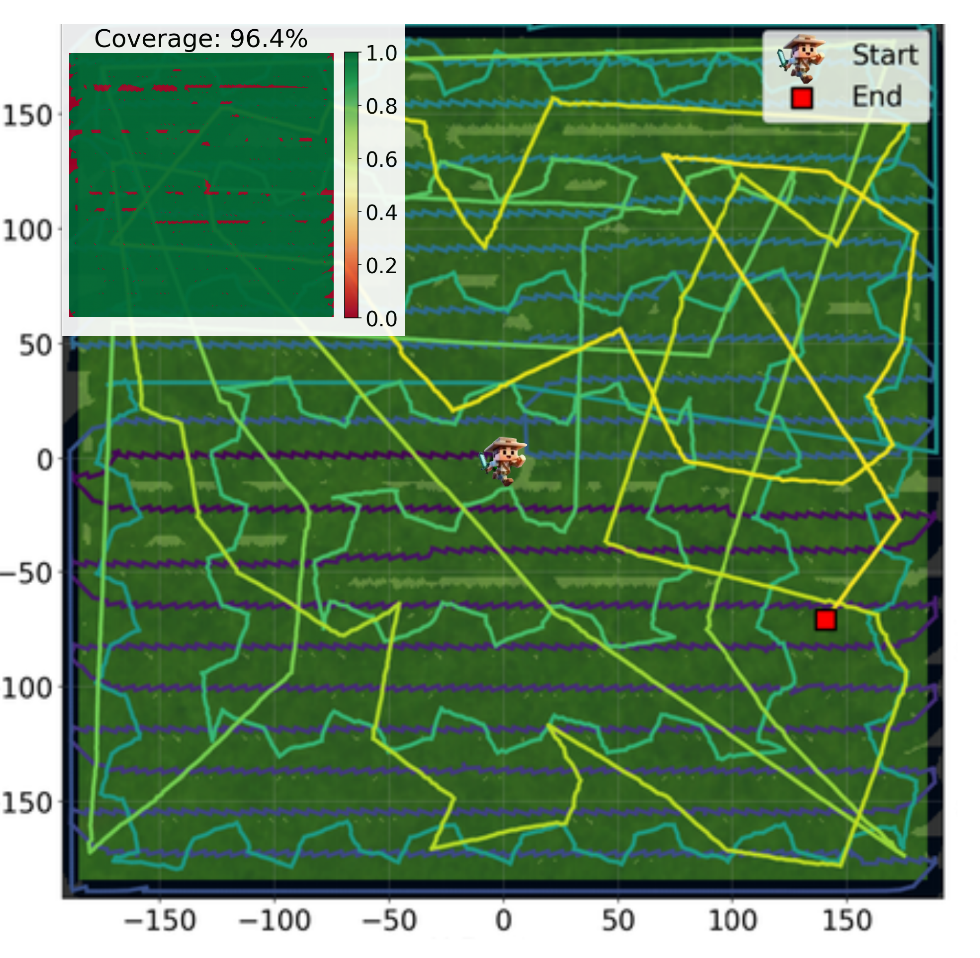}
    \caption{Exploration trajectory comparison on a $384\times384$ simulated map after 10,000 timesteps. Left: MrSteve (75.6\% coverage). Right: WISE (96.4\% coverage). Color indicates temporal progression (blue: early; red: late). MrSteve exhibits locally greedy behavior and repeatedly concentrates exploration near the spawn region, whereas WISE achieves broad and uniform coverage through coordinated global, regional, and local exploration.}
    \label{fig:explore_traj}
\end{figure}

\textbf{Real Environment.}
We next evaluate under the standard Minecraft environment containing stochastic elements including mobs, weather, and day--night cycles. Although environmental uncertainty reduces coverage for all methods, WISE consistently maintains a substantial advantage.
On the $128\times128$ real-world map with a 10,000-step budget, WISE covers 97\% of the environment, compared to 67\% for MrSteve and only 19\% for Steve-1.
The three exploration tiers address complementary failure modes of previous approaches. The global quadtree stage performs  $O(\log n)$ region lookup and next-target selection and avoids local-greedy behavior. This complexity describes spatial-index operations rather than final coverage, which also depends on navigation, terrain, and the timestep budget. The frontier-based regional tier expands exploration boundaries toward unexplored areas. Finally, the Voronoi local completion stage explicitly fills small unvisited regions that remain invisible to higher-level planning.
Trajectory visualizations in Figure~\ref{fig:explore_traj} provide qualitative evidence. MrSteve's count-based exploration tends to concentrate around the spawn location, whereas WISE distributes exploration trajectories more uniformly throughout the environment.

Table~\ref{tab:coverage} further reveals two observations. First, coverage gaps become increasingly pronounced as map scale increases. Although the relative difference narrows from 40 percentage points on small maps to 15 points on large maps, the latter corresponds to substantially larger unexplored regions in absolute area. The smaller normalized gain of the quadtree-only variant on the $384\times384$ map reflects the larger share of the fixed budget spent on long-range travel and local execution, and does not contradict the $O(\log n)$ lookup cost.
Second, gains observed in simulation transfer consistently to realistic environments. On the small real-world map, WISE improves coverage by 30 percentage points over MrSteve and by 78 points over Steve-1, demonstrating strong robustness under environmental stochasticity.

\begin{table}[tb]
\centering
\small
\setlength{\tabcolsep}{7pt}

\caption{Map coverage on 128×128 and 384×384 maps in the simulated and Minecraft environments. Map coverage ($\in[0,1]$) at different Minecraft timestep budgets (20 timesteps $\approx$ 1 second). Columns are grouped by map size. Best results in each column are highlighted in \textbf{bold}. WISE$^{*}$ uses global-level search when exploring.}
\label{tab:coverage}

% ===================== (a) =====================
\begin{tabular*}{\linewidth}{@{\extracolsep{\fill}} l c c c c c c}
\toprule
Method & \multicolumn{3}{c}{$128\times128$} & \multicolumn{3}{c}{$384\times384$} \\
\cmidrule(lr){2-4}\cmidrule(lr){5-7}
& 500 steps & 1000 steps  & 6000 steps  & 1000 steps & 4000 steps & 20000 steps \\
\midrule
MrSteve(Count-Based)                 & 0.46 & 0.57 & 0.59 & 0.17 & 0.53 & 0.83\\
WISE$^{*}$(Quadtree-Based) & 0.65 & \textbf{0.87}  & 0.93 & \textbf{0.19} & 0.72 & 0.83 \\
WISE(ours)              & \textbf{0.66} & 0.85  & \textbf{0.99}  & \textbf{0.19} & \textbf{0.73}  & \textbf{0.98} \\
\bottomrule
\end{tabular*}

% \vspace{2pt}
\begin{tabular*}{\linewidth}{@{\extracolsep{\fill}} p{\linewidth}}
\footnotesize (a) Map Coverage on $128\times128$ and $384\times384$ map in simulated environment.
\end{tabular*}

% \vspace{6pt}

% ===================== (b) =====================
\begin{tabular*}{\linewidth}{@{\extracolsep{\fill}} l c c c c}
\toprule
Method & \multicolumn{2}{c}{$128\times128$} & \multicolumn{2}{c}{$384\times384$} \\
\cmidrule(lr){2-3}\cmidrule(lr){4-5}
& 5{,}000 steps & 10{,}000 steps & 20{,}000 steps & 40{,}000 steps \\
\midrule
Steve-1(Random)                 & 0.11 & 0.19 & 0.06 & 0.13 \\
MrSteve(Count-Based)                   & 0.64 & 0.67 & 0.39 & 0.69 \\
WISE$^{*}$(Quadtree-Based) & \textbf{0.72} & 0.75 & \textbf{0.44} & 0.74 \\
WISE(ours)              & 0.71 & \textbf{0.97} & \textbf{0.44} & \textbf{0.83} \\
\bottomrule
\end{tabular*}

% \vspace{2pt}
\begin{tabular*}{\linewidth}{@{\extracolsep{\fill}} p{\linewidth}}
\footnotesize (b) Map Coverage of different exploration policies in Minecraft.
\end{tabular*}

\end{table}

\subsection{Sequential Sparse Task Performance (ABA-Sparse)}
The ABA-Sparse task evaluates long-term memory and spatial recall under sparse-resource settings. The agent must first collect a sparse resource A, then acquire a dense resource B, and finally return to the original location of A without re-exploration. Each episode has a budget of 12,000 timesteps, and success requires completing all three stages. 
We evaluate each method in two runs of 50 episodes each on the same fixed map layout. The same initialization seeds are used across all methods within each run.

Table~\ref{tab:aba_abc} summarizes the results. Steve-1 fails completely, achieving a 0\% success rate due to the absence of an episodic memory mechanism. MrSteve succeeds in 32\% of episodes with an average completion time of 8,123 steps. WISE$^{*}$(SAEG only), which introduces semantic memory but removes multi-scale exploration and opportunistic scheduling, improves success to 47\% while maintaining a similar completion time (8,093 steps). The full WISE system achieves a success rate of 62\% with an average of only 5,981 steps, corresponding to a 30-point improvement over MrSteve and a 26\% reduction in completion time.

\begin{figure}[htbp]
    \centering

    \begin{subfigure}[b]{0.47\textwidth}
        \centering
        \includegraphics[width=0.93\linewidth]{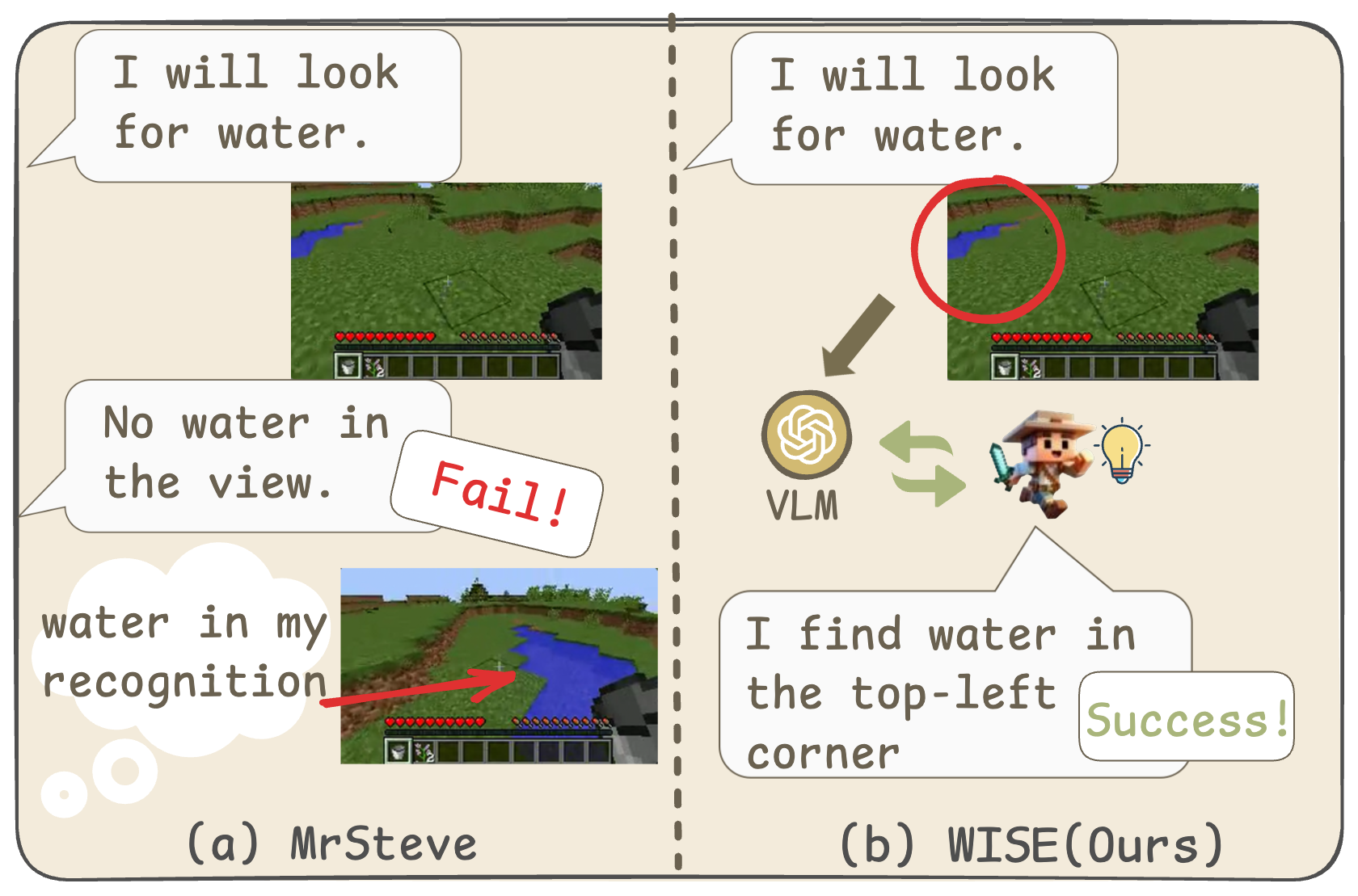}
        \caption{Memory retrieval behaviour in ABA-Sparse Task}
        \label{fig:aba}
    \end{subfigure}
    \hfill % 自动填充间距，使子图左右分开
    \begin{subfigure}[b]{0.50\textwidth}
        \centering
        \includegraphics[width=1\linewidth]{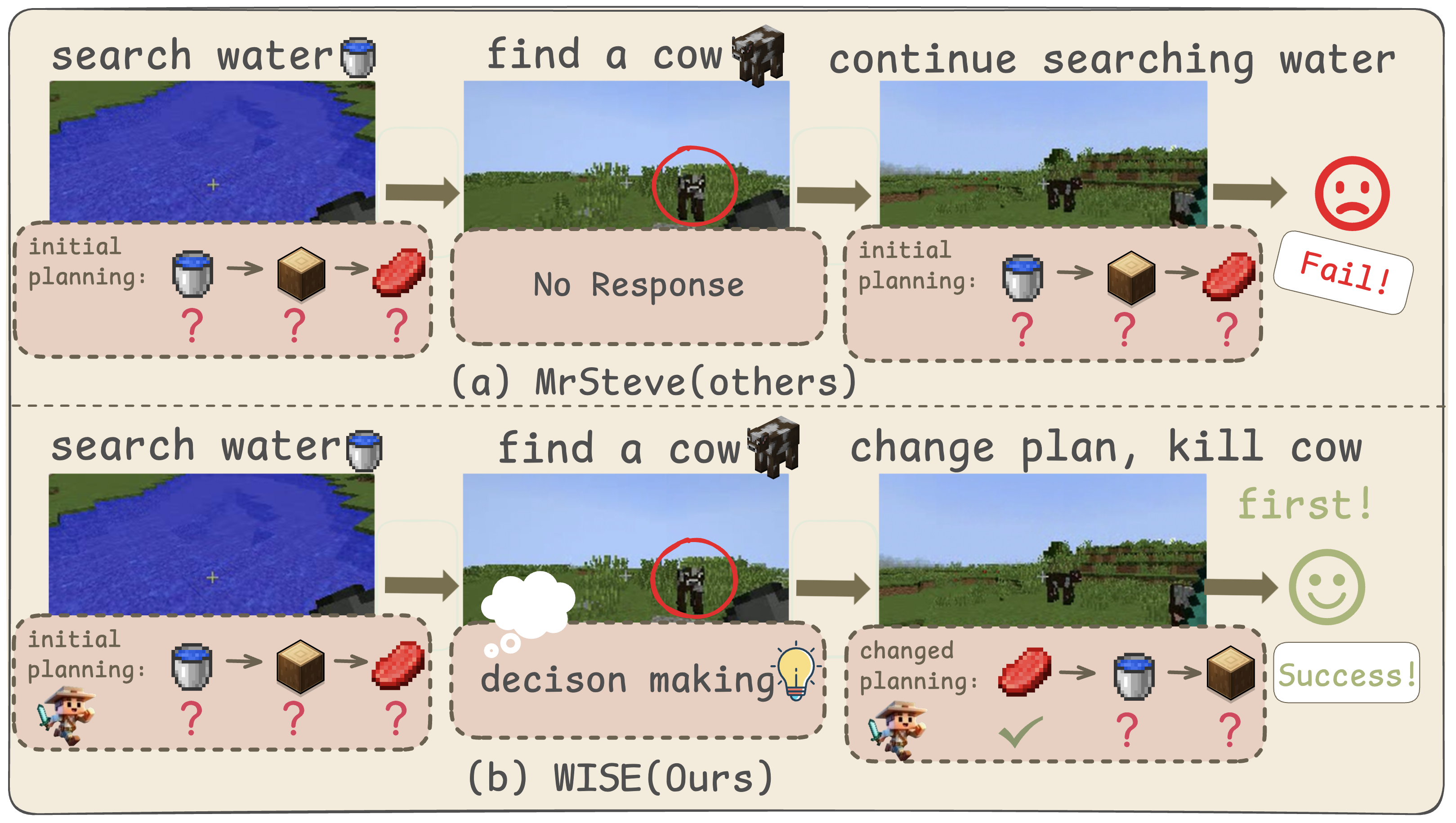}
        \caption{Opportunistic task execution in ABC-Sparse Task.}
        \label{fig:abc}
    \end{subfigure}
    \caption{(a) Memory retrieval in the ABA-Sparse task return phase. MrSteve stores raw visual features; under viewpoint changes or occlusion, cosine similarity between current observations and stored features decreases, leading to retrieval failures and forced re-exploration. In contrast, WISE processes keyframes using a VLM and stores entities and affordance relations in SAEG for retrieval. (b) Opportunistic task execution in the ABC-Sparse task. Each row illustrates the agent's planning state over time. Upon detecting resource C (cow), WISE immediately reorders its subtask queue to prioritize the opportunistic encounter, eliminating a future navigation round-trip. MrSteve's fixed-sequence planner continues executing the original plan and misses the opportunity entirely.}
    \label{fig:compare_trajectory}
\end{figure}

The primary challenge arises during the return-to-A phase. MrSteve's Place Event Memory stores raw visual representations and retrieves memories through feature similarity. As shown in Figure~\ref{fig:aba}, under viewpoint changes, partial occlusion, or environmental variation, cosine similarity frequently drops below the retrieval threshold, causing memory recall failure. Consequently, successful retrieval occurs only in cases where the return trajectory closely matches the original observation viewpoint. WISE addresses this limitation through VLM-generated semantic representations that are  less sensitive to viewpoint changes. The  SAEG-only variant already improves performance by replacing appearance-based matching with semantic retrieval. The full WISE system further incorporates multi-scale exploration, enabling observations of target locations from diverse viewpoints and producing richer memory representations. This results in an additional 15-point gain over  SAEG-only WISE and reduces average return localization time from 4,253 steps (MrSteve) to 2,341 steps.
These results suggest that robust long-term spatial recall in sparse environments requires both semantically grounded memory and sufficiently comprehensive exploration.

\begin{table}[tb]
\centering

\caption{Performance on Sparse Tasks
\\ \small ABA sequential sparse task and ABC non-sequential sparse task: success rate and average completion timesteps over successful episodes.  WISE$^{*}$ (SAEG only) removes multi-scale exploration and
opportunistic scheduling. \textbf{Bold} = best. --- = not applicable.}
\label{tab:aba_abc}

\resizebox{\linewidth}{!}{%
\begin{tabular}{lcccc}
\toprule
Method & \multicolumn{2}{c}{Sequential Sparse Task (ABA-Sparse)} & \multicolumn{2}{c}{Adaptive Non-Sequential Task (ABC-Sparse) } \\
\cmidrule(lr){2-3}\cmidrule(lr){4-5}
& Success Rate(\%) & Avg. Timesteps &Success Rate(\%) & Avg. Timesteps \\
\midrule
Steve-1                 & 0 & --- & 0 & --- \\
MrSteve(PEM)                &32 & 8{,}123 & 33 & 8{,}040 \\
 WISE$^{*}$ (SAEG only) & 47 & 8{,}093 & 42 & 8{,}325 \\
WISE(ours)              & \textbf{62} & \textbf{5{,}981} & \textbf{77} & \textbf{4{,}620} \\
\bottomrule
\end{tabular}%
}
\end{table}

\subsection{Adaptive Non-Sequential Task Performance (ABC-Sparse)}
The ABC-Sparse task evaluates opportunistic decision-making under dynamic task conditions. The agent must collect resources A, B, and C, but the completion order is unconstrained. Resource C may appear while the agent is executing another subtask, requiring dynamic reassessment of task priorities. Fixed-sequence planners are expected to systematically miss such opportunities.

Table~\ref{tab:aba_abc} presents the quantitative results. MrSteve achieves only a 33\% success rate with an average completion time of 8,040 steps. Because its planner follows a rigid execution sequence (water $\rightarrow$ logs $\rightarrow$ beef), the agent typically finishes water and logs before attempting to locate beef, often requiring a return trip exceeding 150 blocks. In many episodes, cows either cannot be found within the step budget or are encountered too late.
WISE$^{*}$(SAEG only), which incorporates semantic memory but lacks opportunistic scheduling, improves success to 42\%. Although semantic memory enables later retrieval of cow locations, the fixed task order still prevents immediate exploitation of encountered opportunities.
The full WISE system achieves a success rate of 77\% with an average completion time of only 4,620 steps, corresponding to a 44-point improvement over MrSteve and a 42.5\% reduction in completion steps.
The improvement arises directly from the Opportunistic Task Scheduler. When a cow enters the agent's observation space,  SAEG provides the semantic affordance relation:
 $cow \rightarrow CAN\_OBTAIN \rightarrow beef$. 
 
 This immediately increases the  semantic affordance relevance score for the ``obtain beef'' subgoal while reducing its navigation cost. As shown in Figure~\ref{fig:abc}, the scheduler recomputes task priorities and dynamically reorders execution to [beef, water, logs]. The agent collects beef immediately and resumes the remaining objectives afterward.
This mechanism eliminates expensive future revisitation and directly explains the observed efficiency gain. Failure cases primarily arise when opportunities cannot be exploited in time---for example, when cows move away or are not detected before leaving the observation range.
These results demonstrate that coupling semantic memory with adaptive scheduling transforms the low-level controller from a passive executor into an active decision-maker capable of exploiting environmental opportunities.

\subsection{Ablation Study}
We conduct controlled ablations to quantify the contribution of individual components within WISE. First, we analyze the effectiveness of the exploration hierarchy on a $128\times128$ map. Second, we remove each major system component--- SAEG, Multi-Scale Progressive Exploration, and Opportunistic Task Scheduler---to investigate their  individual contributions on the ABC-Sparse task. We also add VLM Entity Retrieval without SAEG to isolate graph structure from VLM-generated entity labels. All ablations use the same hyperparameters and evaluation protocol as the full model.

\subsubsection{Exploration Tier Ablation}
We evaluate three exploration variants on a $128\times128$ map: (1) Global-only, which retains only the quadtree-based global exploration tier; (2) Global+Regional, which combines quadtree and frontier-based exploration; and (3) Full WISE, which additionally incorporates the local Voronoi completion stage. Experiments are conducted in the simulated environment to remove low-level movement stochasticity, ensuring that performance differences reflect the intrinsic efficiency of the exploration algorithms.

\begin{table}[tb!]
\centering
\small  % 匹配示例的字号

\caption{Ablation Studies of WISE Method
\\ \small Ablation results. \textbf{Bold} = best. --- = not applicable.}
\label{tab:ablation}

% ===================== (a) 探索层级消融 =====================
\begin{subtable}{\linewidth}
\centering
\begin{tabular*}{\linewidth}{@{\extracolsep{\fill}} p{3.5cm} c c c c}
\toprule
Variant & 5{,}000 steps & 7{,}000 steps & 10{,}000 steps & Cov. Eff. \\
\midrule
MrSteve (Count-Based)                 & 0.64         & 0.65        & 0.67         & 0.067          \\
Global only               & \textbf{0.72} & 0.75        & 0.75         & 0.075          \\
Global + Regional                     & 0.71         & 0.87        & 0.91         & 0.091          \\
Global + Regional + Local (full WISE) & 0.71         & \textbf{0.87} & \textbf{0.97} & \textbf{0.097} \\
\bottomrule
\end{tabular*}
\caption{Exploration tier ablation on $128\times128$ map. Coverage efficiency = coverage per 1,000 timesteps at 10,000-step budget.}
\label{tab:ablation_tier}
\end{subtable}

\vspace{6pt}  % 子表间间距，匹配示例

% ===================== (b) 模块消融 =====================
\begin{subtable}{\linewidth}
\centering
\begin{tabular*}{\linewidth}{@{\extracolsep{\fill}} p{5.8cm} c c c}
\toprule
Variant & Success Rate (\%) & Avg. Timesteps & $\Delta$ vs. full model \\
\midrule
WISE w/o SAEG (feature retrieval)      & 31          & 8{,}034          & $-$46 pts \\
VLM Entity Retrieval without SAEG & 56 & 6{,}320 & $-$21 pts \\
WISE w/o Multi-Scale Exploration & 45          & 7{,}942          & $-$32 pts \\
WISE w/o Opportunistic Task Scheduler     & 62          & 6{,}981          & $-$15 pts \\
WISE (full model)    & \textbf{77} & \textbf{4{,}620} & ---       \\
\bottomrule
\end{tabular*}
\caption{Module ablation on adaptive non-sequential (ABC) task ($128\times128$ real map). $\Delta$ vs.\ full model = success rate difference vs.\ full WISE.}
\label{tab:ablation_module}
\end{subtable}

\end{table} 

Table~\ref{tab:ablation_tier} reports map coverage at 5,000, 7,000, and 10,000 timesteps, together with coverage efficiency, defined as coverage per 1,000 timesteps at the 10,000-step budget. The Global-only variant achieves 75\% coverage at 10,000 steps, already outperforming MrSteve's count-based strategy (67\%). However, coverage rapidly plateaus because large interior regions and boundary-adjacent gaps remain unexplored. Introducing the regional frontier stage increases coverage substantially to 91\%, while the complete three-tier design further improves coverage to 97\%. Coverage efficiency similarly increases from 0.075 (Global-only) to 0.091 (Global+Regional), reaching 0.097 for the full system. These results indicate that each exploration tier contributes complementary gains.

The improvements can be explained by the distinct roles of each tier. The global quadtree stage prevents the agent from becoming trapped in local neighborhoods by encouraging exploration toward coarse-scale and distant regions. However, once a target region is selected, it does not specify where exploration should continue within that region, often leading to inefficient local wandering. The frontier-based regional stage addresses this limitation by explicitly identifying boundaries between explored and unexplored areas and directing the agent toward informative viewpoints. Nevertheless, even with these two stages, isolated gaps may persist---for example, a small unexplored area hidden behind terrain structures. Such regions frequently remain undetected because they do not lie on active frontiers. The local Voronoi stage explicitly addresses this issue by partitioning the explored space and directing the agent toward the largest remaining unexplored regions. Without this final stage, coverage saturates at 91\%; with it, the agent achieves near-complete exploration coverage (97\%).

\subsubsection{Module Ablation}

We further evaluate ablation variants on the ABC-Sparse task using a $128\times128$ real-world map, with the results reported in Table~\ref{tab:ablation_module}.
Specifically, WISE w/o SAEG uses raw feature-based retrieval; VLM Entity Retrieval without SAEG retains GPT-4o-generated entity labels but removes semantic affordance edges and graph traversal; WISE w/o Exploration substitutes the proposed multi-scale exploration strategy with MrSteve's count-based policy; and WISE w/o Opportunistic Task Scheduler adopts a fixed execution order instead of dynamic task prioritization.

\textbf{ Removing SAEG and reverting to feature retrieval} causes the largest performance degradation, reducing success rate from 77\% to 31\%, nearly matching MrSteve's 33\%. Even with efficient exploration and adaptive scheduling, the absence of semantic affordance memory prevents reliable retrieval of resource locations and removes the structured task relevance used by the scheduler.

\textbf{VLM Entity Retrieval without SAEG} reaches 56\% success, showing that VLM labels help recognition while the remaining gap to full WISE reflects the contribution of affordance edges and graph-based retrieval.

\textbf{Removing Multi-Scale Progressive Exploration} reduces success to 45\%, corresponding to a 32-point drop relative to the full model. Although semantic memory and opportunistic planning remain available, the agent explores less effectively and consequently acquires fewer useful observations. Resource encounters become increasingly sparse, and observations may be collected from unfavorable viewpoints that fail to generate robust memory representations. These results suggest that exploration quality fundamentally constrains the information available for downstream memory and planning.

\textbf{Removing the Opportunistic Task Scheduler} produces a smaller but still substantial reduction, decreasing success to 62\%. In this setting, the agent can still discover and remember resource locations, but it can no longer dynamically reorder subtasks. As a result, the agent completes water and log collection before revisiting previously observed cows, introducing unnecessary travel costs. The 15-point gap largely corresponds to scenarios in which opportunities are highly time-sensitive.

The component-removal results follow the ordering:

\[
\begin{aligned}
& \text{SAEG (feature retrieval only)}~(-46)\\
&\quad > \text{Multi-Scale Exploration}~(-32)
> \text{Opportunistic Scheduler}~(-15).
\end{aligned}
\]

This hierarchy closely mirrors the dependency structure of WISE itself: exploration supplies observations, memory transforms observations into semantic knowledge, and planning consumes that knowledge to make adaptive decisions. Notably,  reverting to feature retrieval reduces performance to approximately MrSteve's level. Conversely, memory without advanced exploration still outperforms MrSteve (45\% vs. 33\%), indicating that semantic memory retains value even when observation quality is limited. The complete system, integrating all three modules, achieves 77\% success and substantially outperforms all ablated variants.

\section{Conclusion}
\label{sec:conclusion}
In this paper, we introduced WISE, a low-level controller for long-horizon embodied tasks in Minecraft that unifies exploration, memory, and planning within a closed reasoning loop. WISE augments conventional episodic memory with a  Semantic Affordance Event Graph that captures not only what, where, and when, but also why past observations matter and which future tasks they enable. Based on this semantic memory, an Opportunistic Task Scheduler dynamically re-prioritizes subtasks, while a multi-scale progressive exploration strategy provides spatially diverse observations.
Experiments on sparse sequential and non-sequential tasks demonstrate that WISE substantially outperforms strong baselines in exploration efficiency, task success rate, and completion speed.  Ablation studies further show complementary contributions from exploration, semantic memory, and adaptive planning. Future work will extend WISE to broader embodied environments and investigate lightweight alternatives to external VLMs for scalable knowledge construction.

\section{Limitations}

WISE builds on established techniques, with its main contribution lying in their closed-loop integration through semantic affordance memory, opportunistic scheduling, and multi-scale exploration. Sparse-task evaluation uses a single fixed Minecraft map, while exploration experiments use fixed map instances at two scales; broader cross-world generalization remains to be validated. Future work will investigate jointly learned affordance representations and scheduling policies and extend evaluation across diverse world seeds, biomes, and task families.

\subsubsection*{Acknowledgments}
This work was supported by the National Natural Science Foundation of China under Grant 62573370 and the Education Department of Guangdong Province under Grant 2025ZDZX3051.

% \newpage 
% \input{sec/supple1}

% \bibliography{main}

\bibliography{main}
\bibliographystyle{tmlr}

\appendix
% \section{Appendix}
\newpage 
% ========================================================
% WISE Appendix -- paste after \bibliography{} in main.tex
% Requires: \usepackage{booktabs, multirow, listings, xcolor}
% ========================================================

\appendix

\section{Implementation Details}
\label{app:impl}

\subsection{Multi-Scale Progressive Exploration}
The explorer must supply memory module with observations that are both spatially complete and efficiently acquired. No single-scale strategy can satisfy both requirements simultaneously: global-only strategies leave local gaps; local-only strategies scale poorly; frontier-only strategies ignore interior voids. WISE addresses this with a \textbf{three-tier progressive framework} in which each tier resolves the gaps left by the previous one.

\textbf{\textit{Global-Scale Rapid Search.}}
The global tier answers: which macro-region should the agent prioritise next? The Minecraft world is represented as an \textbf{adaptive quadtree} whose root spans the full known map and whose children represent candidate regions. Subdivision adapts dynamically to visitation frequency, allocating finer resolution to denser unexplored areas. Let $P_t$ denote the agent's current 2D map position. Each candidate node $N_i$ is scored by:
\begin{equation}
Score_G(N_i)=\alpha_G \cdot Depth(N_i)-\beta_G \cdot Dist(P_t,N_i),
\label{eq:quadtree-score}
\end{equation}

where $Depth(N_i)$ measures the normalized depth or unexplored granularity of region $N_i$, $Dist(P_t,N_i)$ is the normalized distance from the current position to the region, and $\alpha_G,\beta_G$ are global-tier weights. The first term encourages the agent to visit fine-grained unexplored regions, while the second term penalizes distant targets. The quadtree reduces the cost of region lookup and next-target selection from $O(n)$ to $O(\log n)$ over candidate spatial regions. This complexity refers to the efficiency of selecting the next macro-region from the spatial index. The realized coverage still depends on the low-level controller, terrain geometry, failed movements, and the timestep budget.

\textbf{\textit{Regional-Scale Frontier Refinement.}}
Given a macro-region from the global tier, the regional tier answers: which boundary points offer the highest information gain? Frontier points---boundaries between explored and unexplored space---are detected via morphological dilation and scored as:
\begin{equation}
Score_R(f_i)=
\alpha_R \cdot Cover(f_i)
+
\beta_R \cdot Novel(f_i)
-
\gamma_R \cdot Dist(P_t,f_i),
\label{eq:frontier-score}
\end{equation}

where $Cover(f_i)$ estimates the uncovered area visible behind frontier $f_i$, $Novel(f_i)$ measures information novelty, $Dist(P_t,f_i)$ is the normalized distance from the agent to the frontier, and $\alpha_R,\beta_R,\gamma_R$ are regional-tier weights. The frontier with the highest score is selected as the next regional exploration target. This tier expands the explored boundary and reduces the local wandering that can occur when only global target selection is used.

\textbf{\textit{Local-Scale Voronoi Completion.}}
After frontier refinement, residual interior patches may remain---a problem that neither global planning nor frontier-following fully resolves. The local tier answers: which residual gaps remain unfilled? We apply Voronoi decomposition\cite{krozel1990navigation} over the visited point set $\mathbf{V} = \{v_1, v_2, \dots, v_n\}$, where $v_i = (x_i, y_i, t_i)$, $x_i$ and $y_i$ are 2D coordinates, and $t_i$ is the timestamp. The Voronoi cell of visited point $v_i$ is
\begin{equation}
\mathbf{C_i} = \{ p \in \boldsymbol{\Omega} \mid \|p - v_i\|_2 < \|p - v_j\|_2,\ \forall j \neq i \}.
\label{eq:voronoi_cell}
\end{equation}
Residual uncovered regions are identified from $\boldsymbol{\Omega} \setminus \bigcup_{i=1}^n \mathbf{C_i}$. Each residual region $U_i$ is prioritised by:
\begin{equation}
Score_L(U_i)=
\alpha_L \cdot Area(U_i)
-
\beta_L \cdot Dist(P_t,U_i),
\label{eq:voronoi-score}
\end{equation}

where $Area(U_i)$ is the normalized area of the residual unexplored region, $Dist(P_t,U_i)$ is the normalized distance from the current position to that region, and $\alpha_L,\beta_L$ are local-tier weights. The agent moves toward the highest-scoring residual region to improve local coverage completion.

This local tier improves coverage completion by explicitly targeting residual gaps that remain after global and regional exploration. Together, the three tiers provide a coarse-to-fine exploration strategy: the global tier selects macro-regions efficiently, the regional tier expands frontier boundaries, and the local tier fills remaining uncovered regions.

\textbf{\textit{Local escape mechanism.}}
Three escalating recovery heuristics prevent exploration stagnation. If fewer than five blocks are traversed in 30 seconds, the agent triggers \textit{local replanning}. On repeated failure to reach a target, it performs a \textit{target reset} via memory query. On repeated location revisitation, it executes a \textit{graph-level rollback} to the nearest unvisited key node. These heuristics ensure that exploration continues to supply fresh observations to the memory module even under adverse terrain conditions.

\subsection{Asynchronous VLM Update and SAEG Maintenance}
\label{app:async_vlm}

WISE decouples VLM inference from the online control loop. During execution, the online controller selects candidate keyframes from the short-term geometric memory and appends them to a pending VLM buffer. An asynchronous VLM worker processes the buffer when 16 keyframes are available or flushes it after 25 seconds. While the request is being processed, the agent continues acting using $M_{\mathrm{short}}$ and the latest available version of $M_{\mathrm{SAEG}}$.

Once a VLM response is returned, the parsed outputs are grounded against the Minecraft task-resource vocabulary and incorporated into SAEG by updating entity nodes, spatial attributes, timestamps, semantic affordance relations, and co-occurrence relations. This design prevents VLM latency from blocking real-time action execution while keeping the graph available for semantic affordance retrieval and opportunistic scheduling.

The number of selected keyframes is adaptive and depends on the explored trajectory, entropy filtering, clustering, and redundancy removal. A typical $128\times128$ task yields approximately 200--300 processed keyframes. We report the average number of VLM requests per episode in Section~\ref{app:cost}.

\subsection{Hardware and Software Configuration}
\label{app:hardware}

All experiments were conducted on a single server equipped with four NVIDIA RTX 4080 16GB GPUs. Unless otherwise stated, each agent rollout used one GPU. The host operating system was Ubuntu v22.04. All agents were implemented in Python v3.11 with PyTorch v2.5.1.

Minecraft Java Edition was accessed via \href{https://docs.minedojo.org/sections/getting_started/install.html}{MineDojo v0.1 (built on MineRL)}.
The low-level controller used the publicly released
\href{https://openaipublic.blob.core.windows.net/minecraft-rl/models/2x.model}{VPT 2x checkpoint}
and the VPT-Nav weights distributed with MrSteve.
GPT-4o requests used the model identifier

\texttt{gpt-4o-2024-05-13}.
Each experiment episode ran with a per-timestep time constraint of $\leq$50\,ms,
consistent with the Minecraft Java Edition real-time rendering requirement.

%----------------------------------------------------------
\subsection{World and Map Configuration}
\label{app:world}

\begin{figure}[!ht]
    \centering
    \includegraphics[width=0.7\linewidth]{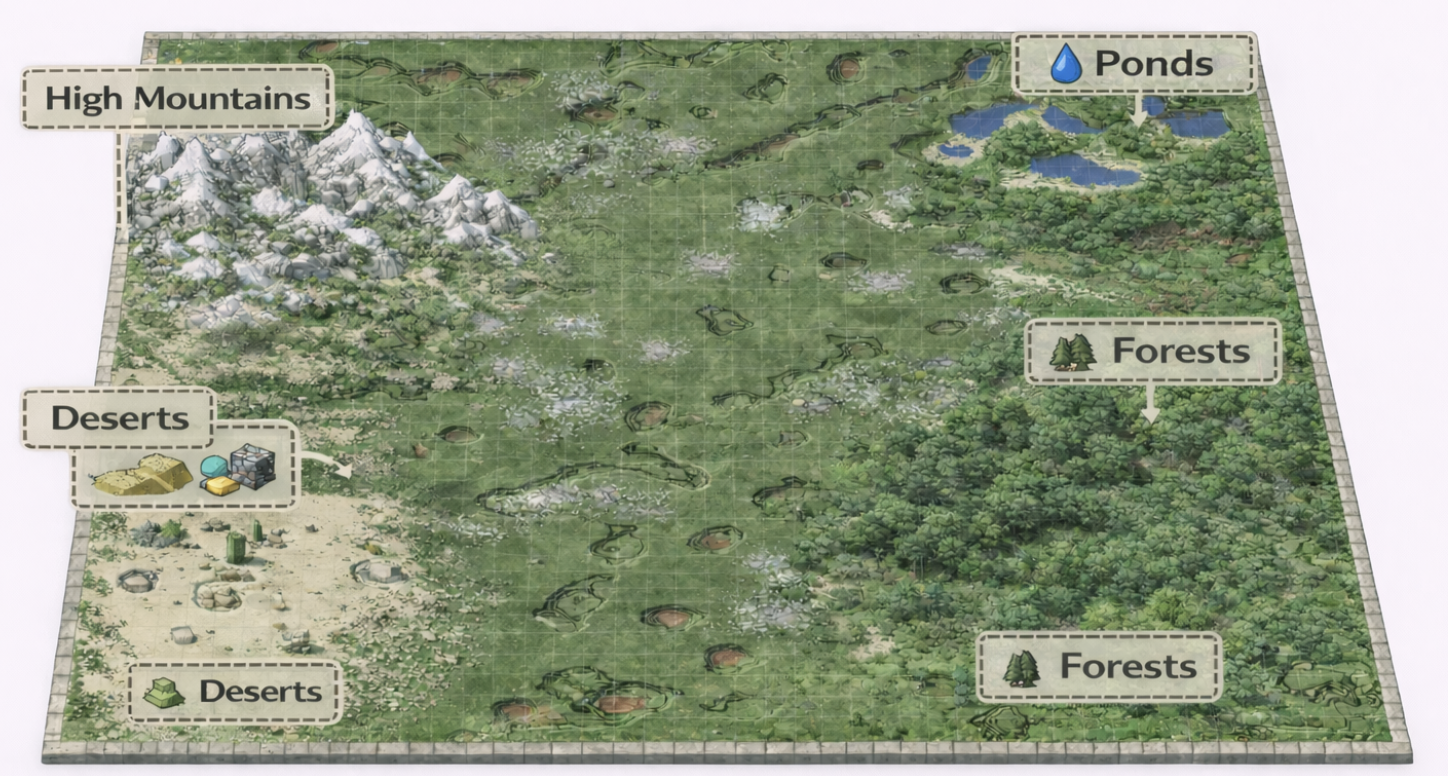}
    \caption{The 128$\times$128 real Minecraft map used for task-completion experiments. The combination of biome diversity and deliberate resource sparsity requires WISE's memory, scheduling, and exploration modules to operate jointly.}
    \label{fig:task_map}
\end{figure}

\paragraph{Sparse task experiments (Sections~4.3 and~4.4).}
A $128\times128$ real Minecraft map combining mountains, ponds, desert, and forest areas was used as shown in Figure~\ref{fig:task_map}.

Each method was evaluated in two runs of 50 episodes each on the same fixed map layout.  Episodes used different random initialization seeds, while all methods were evaluated using the same set of seeds, spawn-point protocol, and episode budget.

\paragraph{Exploration experiments (Section~4.2).}
Two map scales were used: small ($128\times128$ blocks) and large
($384\times384$ blocks), each evaluated under two conditions.
\begin{itemize}
    \item \textbf{Simulated}: mobs disabled, terrain variation disabled,
          day--night cycle paused.
    \item \textbf{Real (Stochastic)}: full Minecraft stochasticity including
          hostile mobs, weather, and day--night cycles.
\end{itemize}
All agents started from the same fixed spawn point on the same world seed
within each condition.

All methods are evaluated on identical map instances and seeds within each condition.

%----------------------------------------------------------
\subsection{Evaluation Protocol and Coverage Metric}
\label{app:eval_protocol}

Map coverage is computed as the number of unique map cells visited or observed within the agent's field of view, divided by the total number of map cells in the evaluation region. Coverage is accumulated over the episode, with repeated observations of the same cell counted only once.
A timestep corresponds to a Minecraft environment tick; Minecraft runs at 20 timesteps per second. Coverage values are normalized by map area and are therefore comparable within the same map size and evaluation condition.

For the real-environment coverage table in the main paper, the 5,000- and 10,000-timestep columns correspond to the $128\times128$ map, while the 20,000- and 40,000-timestep columns correspond to the $384\times384$ map. Therefore, these columns should not be interpreted as a single monotonic sequence on the same map. The columns are grouped by map size to make this correspondence explicit.

For task-completion metrics, success rate is computed over all evaluation episodes. Average completion timesteps are computed over successful episodes only.

%----------------------------------------------------------
\subsection{Baseline Implementation}
\label{app:baseline}

\textbf{Steve-1}: the publicly released checkpoint
was used without modification; it serves as a no-memory reference baseline.
\textbf{MrSteve}: the publicly released checkpoint and
Place--Event Memory (PEM) implementation were used; it serves as the primary
baseline.
All baselines share identical map configurations, spawn points, and episode
budgets with WISE.

MrSteve uses PEM-style visual-similarity memory retrieval and does not use GPT-4o for semantic graph construction. WISE uses GPT-4o only for constructing and updating SAEG.

%----------------------------------------------------------
\subsection{Full Hyperparameter Table}
\label{app:hparam}

Table 4 lists all secondary hyperparameters required for reproduction.

\begin{table}[h]
\centering
\caption{Full hyperparameter list for WISE.}
\label{tab:full_hparam}
\footnotesize
\setlength{\tabcolsep}{4pt}
\begin{tabular}{p{1.4cm}p{5.4cm}p{1.6cm}p{5.6cm}}
\toprule
\textbf{Module} & \textbf{Parameter} & \textbf{Value} & \textbf{Description} \\
\midrule
{Exploration}
  & Quadtree depth weight $\alpha_G$      & 0.6        & Rewards unexplored depth (Eq.\,\ref{eq:quadtree-score})\\
  & Quadtree distance penalty $\beta_G$   & 0.4        & Penalizes navigation cost (Eq.\,\ref{eq:quadtree-score})\\
  & Frontier weights $(\alpha_R,\beta_R,\gamma_R)$ & norm. & Coverage, novelty, distance (Eq.\,\ref{eq:frontier-score})\\
  & Local completion weights $(\alpha_L,\beta_L)$    & norm.      & Area, distance (Eq.\,\ref{eq:voronoi-score})\\
  & Stagnation threshold                & 5 blk/30 s & Triggers local replanning \\
  & Sliding window size                 & {60s} & Short-term memory eviction \\
\midrule
{Memory}
  & Entropy threshold $\theta_{\mathrm{ent}}$        & 0.15       & Entropy-based keyframe trigger \\
  & Cosine threshold $\theta_c$         & 0.85       & Cluster-based trigger \\
  & Grounding threshold $\theta_{\mathrm{ground}}$ & 0.9& Vocabulary/entry matching \\
  & Top-$k$ retrieval                   & 10 & Short-term candidates \\
  & Retrieval weight $\lambda_{\mathrm{ret}}$         & 0.5& Similarity vs.\ semantic affordance matching \\
  & VLM batch size                      & 16 frames & Buffered keyframes per streaming VLM request \\
  & VLM requests per episode               & $\sim$14   & Average across episodes \\
\midrule
{Planning}
  & Urgency weight $w_u$                & 0.3        & Eq.\,\ref{eq:priority}\\
  & Semantic affordance relevance weight $w_a$ & 0.5        & Eq.\,\ref{eq:priority}\\
  & Navigation cost weight $w_n$        & 0.2        & Eq.\,\ref{eq:priority}\\
\bottomrule
\end{tabular}
\end{table}

\noindent
\textbf{Hyperparameter selection.}
Keyframe thresholds ($\theta_{\mathrm{ent}}$, $\theta_c$) were selected via grid search on
a held-out validation map.
Search ranges:
\[
\theta_{\mathrm{ent}}\in\{0.05,0.10,0.15,0.20\},\quad
\theta_c\in\{0.75,0.80,0.85,0.90\}.
\]

Scheduler weights $(w_u,w_a,w_n)$ are reported as the default setting used in the main experiments. Sensitivity analyses are reported in Section~\ref{app:sensitivity}.

%----------------------------------------------------------
\subsection{Semantic Affordance Graph Construction Details}
\label{app:kg}

The structured Minecraft knowledge database used for grounding VLM outputs was derived from Minecraft Wiki resources and covers mob drops (e.g., cow\,$\to$\,beef), crafting recipes, smelting chains, and environmental state transitions. The database contains 1534 entries and is used to validate candidate entity--outcome relations proposed by the VLM.

\noindent
Graph schema:
\begin{itemize}
    \item \textbf{Node types:} entity (e.g., cow, stone), action (e.g., mine,
          craft), environment (e.g., rain, hunger).
    \item 
          \textbf{Edge types:} timestep, position, semantic affordance (e.g., \texttt{CAN\_OBTAIN}, \texttt{CAN\_CRAFT}, \texttt{CAN\_SMELT}), and co-occurrence (e.g., \texttt{CO\_OCCURS\_WITH}).
\end{itemize}

\noindent

Average number of graph nodes and edges per episode at termination: approximately 120.

\subsection{VLM Prompt and Output Schema}
\label{app:prompt_schema}

Each VLM request receives selected keyframes, timestamps, and agent poses. The VLM returns visible entities, candidate semantic affordance relations, and spatial co-occurrences using the following schema:

\begingroup
\small
\begin{quote}
\ttfamily
Return JSON only. For each keyframe, list visible Minecraft entities, candidate affordance relations from \{CAN\_OBTAIN, CAN\_CRAFT, CAN\_SMELT\}, and spatial co-occurrences:
\{``frames'': [\{``timestep'': int, ``entities'': [...], ``affordances'': [\{``source'': str, ``relation'': str, ``target'': str\}], ``co\_occurrences'': [...]\}]\}.
\end{quote}
\endgroup

After receiving the VLM output, WISE applies four filtering steps: (1) normalize entity names to the Minecraft vocabulary; (2) discard entities or outcomes not present in the task-resource database; (3) merge duplicate entity instances with the same normalized type and nearby spatial coordinates; and (4) retain semantic affordance edges only when the source entity is observed and the target resource is valid under Minecraft mechanics.

%----------------------------------------------------------
\subsection{API Cost Breakdown}
\label{app:cost}

All GPT-4o requests were processed asynchronously and did not block the online control loop. Table~\ref{tab:api_cost} summarizes the API cost.

\begin{table}[!htb]
\centering
\caption{GPT-4o API cost breakdown.}
\label{tab:api_cost}
\small
\begin{tabular}{ll}
\toprule
\textbf{Item} & \textbf{Value} \\
\midrule
 Avg.\ VLM requests per episode       & $\sim$14 \\
Avg.\ input tokens per request       & 4575 \\
Avg.\ output tokens per request      & 1618 \\
Est.\ cost per episode (USD)      & \$0.48\\
Total API cost across all exps.\ (USD) & \$512\\
\bottomrule
\end{tabular}
\end{table}

% ========================================================
% ========================================================
\section{Additional Experimental Analyses}
\label{app:additional_analyses}

% --------------------------------------------------------
\subsection{SAEG Quality}
\label{app:saeg_quality}

To evaluate the quality of the Semantic Affordance Event Graph (SAEG), we
sample 50 keyframes from ABC-Sparse episodes and manually annotate the
task-relevant entities and semantic affordance relations in each frame.
Entity precision is computed as the proportion of predicted entities that
match the manual annotations. Affordance-relation precision is computed as
the proportion of predicted entity--relation--outcome tuples that agree
with both the manual annotations and the curated Minecraft knowledge base.

\begin{table}[H]
\centering
\small
\caption{SAEG quality on 50 manually annotated ABC-Sparse keyframes.}
\label{tab:saeg_quality}
\begin{tabular}{lcc}
\toprule
Representation & Entity precision & Affordance-relation precision \\
\midrule
SAEG & 87.5\% & 87.5\% \\
\bottomrule
\end{tabular}
\end{table}

As shown in Table~\ref{tab:saeg_quality}, most predicted entities and
affordance relations are consistent with the annotations. The remaining
errors mainly arise from small or partially occluded objects and from
ambiguous visual evidence. These results support the use of SAEG for
task-relevant memory retrieval, while the evaluation should be interpreted
as a focused diagnostic analysis because it uses a limited set of
50 keyframes.

% --------------------------------------------------------
\subsection{Controlled Retrieval under Viewpoint and Occlusion Changes}
\label{app:controlled_retrieval}

We conduct a controlled retrieval experiment using a fixed Minecraft scene
containing a pond. The target and scene layout remain unchanged, while the
viewpoint, distance, and degree of partial occlusion are varied across six
observations. For each observation, we test whether MineCLIP feature
retrieval and GPT-4o semantic entity retrieval recover the stored pond
memory. A value of 1 denotes successful retrieval.

\begin{table}[H]
\centering
\small
\setlength{\tabcolsep}{5pt}
\caption{Controlled pond retrieval under different views. Column headers
report the visible fraction of the target; 1 denotes successful retrieval.}
\label{tab:controlled_retrieval}
\begin{tabular}{lcccccc}
\toprule
Method & 3.6\% & 10.2\% & 12.4\% & 17.3\% & 20.8\% & 49.4\% \\
\midrule
MineCLIP feature retrieval
& 0 & 0 & 0 & 0 & 1 & 1 \\
GPT-4o semantic entity retrieval
& 0 & 1 & 1 & 1 & 1 & 1 \\
\bottomrule
\end{tabular}
\end{table}

Table~\ref{tab:controlled_retrieval} shows that MineCLIP feature retrieval
succeeds in two of the six views, whereas GPT-4o semantic entity retrieval succeeds in five of the six views. Feature retrieval is sensitive to changes in low-level
appearance, particularly when the target is small or partially occluded.
Semantic entity retrieval can remain successful when sufficient contextual
evidence is available, even if the current image differs from the stored
view.

The visibility fraction alone does not determine retrieval difficulty:
columns with similar visible fractions may correspond to different
viewpoints, distances, or occlusion patterns. Therefore, this experiment
supports greater robustness under the evaluated changes, but it does not
claim complete invariance to viewpoint or occlusion.

% --------------------------------------------------------
\subsection{Hyperparameter Sensitivity}
\label{app:sensitivity}

We evaluate parameters associated with VLM processing, keyframe selection,
multi-scale exploration, and opportunistic scheduling. These experiments
are included in the appendix because they support implementation choices
but are not central to the main comparison. Default settings are shown in
bold.

\subsubsection{VLM Batching and Keyframe Selection}

\begin{table}[H]
\centering
\small
\caption{Sensitivity of VLM batching and keyframe selection.}
\label{tab:memory_sensitivity}
\begin{tabular}{lccc}
\toprule
VLM batch size & 12 & \textbf{16} & 20 \\
\midrule
Mean response time (s) & 39.0 & 44.1 & 51.1 \\
Mean requests per episode & 19 & 14 & 11 \\
\bottomrule
\end{tabular}

\vspace{6pt}

\begin{tabular}{lcccc}
\toprule
$(\theta_c,\theta_{\mathrm{ent}})$
& $(0.65,0.35)$
& $(0.75,0.25)$
& $\mathbf{(0.85,0.15)}$
& $(0.95,0.05)$ \\
\midrule
Selected keyframes & 357 & 281 & \textbf{217} & 425 \\
\bottomrule
\end{tabular}
\end{table}

Table~\ref{tab:memory_sensitivity} illustrates the trade-off introduced by
VLM batching. A smaller batch reduces the response time but produces more
requests per episode, while a larger batch reduces the number of requests
at the cost of increased response time. The default batch size of 16
provides an intermediate operating point between update frequency and API
efficiency.

The keyframe thresholds jointly control visual redundancy and semantic
informativeness. The default setting selects 217 keyframes, reducing VLM
processing relative to the less selective configurations while retaining
the observations used to construct SAEG. Because the two thresholds affect
different selection criteria, the number of selected keyframes is not
expected to vary monotonically when both are changed simultaneously.

\subsubsection{Exploration and Scheduler Parameters}

\begin{table}[H]
\centering
\small
\caption{Sensitivity of the quadtree and scheduler weights.}
\label{tab:planning_sensitivity}
\begin{tabular}{lccc}
\toprule
Quadtree depth weight $\alpha_G$
& 0.48 & \textbf{0.60} & 0.72 \\
\midrule
Coverage at 5,000 timesteps
& 0.56 & \textbf{0.71} & 0.52 \\
Coverage at 10,000 timesteps
& 0.86 & \textbf{0.97} & 0.84 \\
\bottomrule
\end{tabular}

\vspace{6pt}

\begin{tabular}{lcccc}
\toprule
$(w_u,w_a,w_n)$
& $\mathbf{(0.3,0.5,0.2)}$
& $(0.5,0.3,0.2)$
& $(0.3,0.2,0.5)$
& $(0.3,0.4,0.3)$ \\
\midrule
Success rate (\%)
& \textbf{77} & 68 & 56 & 74 \\
Avg.\ completion timesteps
& \textbf{4,620} & 5,830 & 6,650 & 5,225 \\
\bottomrule
\end{tabular}
\end{table}

For global exploration, a small $\alpha_G$ underweights unexplored
fine-grained regions, whereas an excessively large value may prioritize
region depth over navigation efficiency. The default value
$\alpha_G=0.60$ achieves the highest coverage at both evaluated budgets.

The scheduler results show that assigning the largest weight to semantic
affordance relevance, while retaining urgency and navigation-cost terms,
produces the highest success rate and the lowest average completion time
among the evaluated settings. Increasing the urgency weight makes the
scheduler behave more like the original fixed task order, whereas a large
navigation-cost weight can favor nearby but less task-relevant targets.
These results support the selected default configuration but do not imply
that it is optimal for every task distribution.

% --------------------------------------------------------
\subsection{Longer and More Diverse Sparse-Task Sequence}
\label{app:mixed5}

The main ABC-Sparse experiment contains three objectives. To examine a
longer and more heterogeneous sequence, we additionally evaluate the
Mixed-5 Sparse Task:
\[
\texttt{obtain beef}
\rightarrow \texttt{obtain wheat seeds}
\rightarrow \texttt{collect logs}
\rightarrow \texttt{obtain wool}
\rightarrow \texttt{find water}.
\]
The sequence includes mobs, vegetation, collectible resources, and an
environmental landmark, thereby requiring the agent to retain and retrieve
different types of task-relevant observations.

\begin{table}[H]
\centering
\small
\caption{Success rate on the Mixed-5 Sparse Task.}
\label{tab:mixed5}
\begin{tabular}{lc}
\toprule
Method & Success rate \\
\midrule
Steve-1 & 0\% \\
MrSteve (PEM) & 16\% \\
WISE & \textbf{56\%} \\
\bottomrule
\end{tabular}
\end{table}

As shown in Table~\ref{tab:mixed5}, Steve-1 does not complete the full
sequence, while MrSteve reaches a 16\% success rate. WISE achieves 56\%,
indicating that its advantage is retained when the symbolic task sequence
is extended from three to five heterogeneous objectives. This experiment
tests a longer task composition within the same Minecraft domain; broader
generalization to other environments remains outside the present
evaluation.

% ========================================================
\section{Task Definitions}
\label{app:tasks}

\subsection{ABA-Sparse (Sequential Task)}
\label{app:aba}

The agent spawns in the $128\times128$ real Minecraft map.
The task proceeds in three phases:
\begin{enumerate}
    \item \textbf{Phase~$A_1$}: Locate and collect resource~$A$ at
          position~$p_A$.
    \item \textbf{Phase~$B$}: Navigate to and collect distant resource~$B$
          
          (minimum distance from~$A$: $\geq$100 blocks).
    \item \textbf{Phase~$A_2$}: Return precisely to position~$p_A$ using
          only memory---re-exploration is discouraged by the remaining
          timestep budget.
\end{enumerate}

\noindent
A trial is successful if all three phases are completed in the prescribed order within the 12,000-timestep budget.

%----------------------------------------------------------
\subsection{ABC-Sparse (Non-Sequential Task)}
\label{app:abc}

The agent must collect items~$A$, $B$, and~$C$ in \emph{any} order.
Resource~$C$ (e.g., beef from a cow) appears \emph{opportunistically}
mid-episode at a random location during execution of sub-task~$A$,
requiring priority reassessment.

A fixed-sequence policy may miss this opportunity and incur an additional round-trip cost.

\noindent

Specific resource types: \textbf{$A$\,=\,water, $B$\,=\,logs, $C$\,=\,beef (from cow)}.

Timestep budget: \textbf{12,000}.
$C$ spawn distribution: uniform random
within 30 blocks of the agent at timestep 500.

% Publication-style appendix snippet for WISE manuscript.
% Required packages: \usepackage{booktabs,array}
% INTERNAL WORKING DRAFT: replace the success-count and SD macros below
% with values computed from the final episode-level rollout logs.

% ---------- Values to replace after final evaluation ----------
\newcommand{\TTwoMrSteveABASucc}{16}
\newcommand{\TTwoMrSteveABASD}{2050}
\newcommand{\TTwoMrSteveABCSucc}{17}
\newcommand{\TTwoMrSteveABCSD}{2200}
\newcommand{\TTwoSAEGABASucc}{24}
\newcommand{\TTwoSAEGABASD}{1850}
\newcommand{\TTwoSAEGABCSucc}{21}
\newcommand{\TTwoSAEGABCSD}{2050}
\newcommand{\TTwoWISEABASucc}{31}
\newcommand{\TTwoWISEABASD}{1500}
\newcommand{\TTwoWISEABCSucc}{39}
\newcommand{\TTwoWISEABCSD}{1250}

\newcommand{\TThreeNoSAEGSucc}{16}
\newcommand{\TThreeNoSAEGSD}{2250}
\newcommand{\TThreeEntitySucc}{28}
\newcommand{\TThreeEntitySD}{1650}
\newcommand{\TThreeNoExploreSucc}{23}
\newcommand{\TThreeNoExploreSD}{2300}
\newcommand{\TThreeNoSchedulerSucc}{31}
\newcommand{\TThreeNoSchedulerSD}{1850}
\newcommand{\TThreeFullSucc}{39}
\newcommand{\TThreeFullSD}{1250}
% -------------------------------------------------------------

\section{Statistical Details for Sparse-Task Results}
\label{app:statistical_details}

For each method, we conduct two evaluation runs, with 50 episodes
per run. Success rates are averaged across the two runs. Completion
timesteps are reported as mean $\pm$ standard deviation (SD) over
successful episodes pooled across both runs.

\subsection{Main Sparse-Task Results}

\begin{table}[ht!]
\centering
\caption{Run-level success counts and pooled completion-time
statistics for the sparse-task experiments in Table~2.}
\label{tab:sparse_task_statistics}
\small
\setlength{\tabcolsep}{5pt}
\renewcommand{\arraystretch}{1.08}

\begin{tabularx}{\textwidth}{@{}Xcccc@{}}
\toprule
Method
& \shortstack{Run 1\\successes}
& \shortstack{Run 2\\successes}
& \shortstack{Mean success\\rate (\%)}
& \shortstack{Completion timesteps\\(mean $\pm$ SD)} \\
\midrule

\multicolumn{5}{@{}l}{\textit{Sequential Sparse Task (ABA-Sparse)}} \\
Steve-1                  & 0/50  & 0/50  & 0\%  & --- \\
MrSteve (PEM)            & 16/50 & 16/50 & 32\% & 8,123 ± 487\\
WISE$^{*}$ (SAEG only)   & 23/50 & 24/50 & 47\% & 8,093 ± 846 \\
WISE (ours)              & 31/50 & 31/50 & \textbf{62\%}
                        & \textbf{5,981 ± 637} \\

\midrule
\multicolumn{5}{@{}l}{\textit{Adaptive Non-Sequential Task
(ABC-Sparse)}} \\
Steve-1                  & 0/50  & 0/50  & 0\%  & --- \\
MrSteve (PEM)            & 16/50 & 17/50 & 33\% & 8,040 ± 983 \\
WISE$^{*}$ (SAEG only)   & 21/50 & 21/50 & 42\% & 8,325 ± 731 \\
WISE (ours)              & 38/50 & 39/50 & \textbf{77\%}
                        & \textbf{4,620 ± 493} \\
\bottomrule
\end{tabularx}

\vspace{2pt}
\begin{minipage}{\textwidth}
\footnotesize
\textit{Note.} Each run contains 50 episodes. Completion-time
statistics are computed over successful episodes pooled across
the two runs.
\end{minipage}
\end{table}
\subsection{Module Ablation Results}

\begin{table}[ht!]
\centering
\caption{Run-level success counts and pooled completion-time
statistics for the ABC-Sparse ablation in Table~3(b).}
\label{tab:ablation_statistics}
\small
\setlength{\tabcolsep}{4pt}
\renewcommand{\arraystretch}{1.08}

\begin{tabularx}{\textwidth}{@{}Xccccc@{}}
\toprule
Variant
& \shortstack{Run 1\\successes}
& \shortstack{Run 2\\successes}
& \shortstack{Mean success\\rate (\%)}
& \shortstack{Completion timesteps\\(mean $\pm$ SD)}
& \shortstack{$\Delta$ success\\vs. full (pp)} \\
\midrule
WISE w/o SAEG (feature retrieval)
& 15/50 & 16/50 & 31\% & 8,034 ± 421 & $-46$ \\

VLM Entity Retrieval without SAEG
& 28/50 & 28/50 & 56\% & 6,320 ± 767 & $-21$ \\

WISE w/o Multi-Scale Exploration
& 22/50 & 23/50 & 45\% & 7,942 ± 513 & $-32$ \\

WISE w/o Opportunistic Task Scheduler
& 31/50 & 31/50 & 62\% &6,981 ± 608 & $-15$ \\

WISE (full model)
& 38/50 & 39/50 & \textbf{77\%}
& \textbf{4,620 ± 493} & --- \\
\bottomrule
\end{tabularx}
\end{table}


\end{document}